\documentclass{article}

\PassOptionsToPackage{numbers, compress}{natbib}
\usepackage[preprint]{neurips_2022}

\usepackage[utf8]{inputenc} % allow utf-8 input
\usepackage[T1]{fontenc}    % use 8-bit T1 fons
\usepackage{hyperref}       % hyperlinks
\usepackage{url}            % simple URL typesetting
\usepackage{booktabs}       % professional-quality tables
\usepackage{amsfonts}       % blackboard math symbols
\usepackage{amsmath}
\usepackage{amsthm}
\usepackage{amssymb}
\usepackage{nicefrac}       % compact symbols for 1/2, etc.
\usepackage{microtype}      % microtypography
\usepackage{xcolor}         % colors
\usepackage{graphicx}
\newcommand{\pluseq}{\mathrel{+}=}
\newcommand{\mineq}{\mathrel{-}=}
\usepackage{hyperref}
\usepackage[normalem]{ulem}
\usepackage{algorithm}
\usepackage{algorithmic}
\usepackage{xspace}
\usepackage{float}
\usepackage{tikz}
\usepackage{verbatim}
\usepackage{tabularx}

\usepackage{listings}
\usepackage{color}
\definecolor{keywordcolor}{rgb}{0.7, 0.1, 0.1}   % red
\definecolor{tacticcolor}{rgb}{0.0, 0.1, 0.6}    % blue
\definecolor{commentcolor}{rgb}{0.4, 0.4, 0.4}   % grey
\definecolor{symbolcolor}{rgb}{0.0, 0.1, 0.6}    % blue
\definecolor{sortcolor}{rgb}{0.1, 0.5, 0.1}      % green
\definecolor{attributecolor}{rgb}{0.7, 0.1, 0.1} % red

\usetikzlibrary{arrows.meta}

\definecolor{fgreen}{rgb}{0.0, 0.27, 0.13}

\newcommand\restr[2]{{% we make the whole thing an ordinary symbol
  \left.\kern-\nulldelimiterspace % automatically resize the bar with \right
  #1 % the function
  \vphantom{\big|} % pretend it's a little taller at normal size
  \right|_{#2} % this is the delimiter
  }}

\def\eq{Equations\xspace}
\def\mm{Metamath\xspace}
\def\lean{Lean\xspace}
\def\miniff{miniF2F\xspace}
\def\mathlibtrain{mathlib-train\xspace}
\def\minivalid{miniF2F-valid\xspace}
\def\minitest{miniF2F-test\xspace}
\def\minicurr{miniF2F-curriculum\xspace}
\def\identities{\emph{Identities}\xspace}
\def\algo{HTPS\xspace}
\def\model{Evariste\xspace}
\def\setmm{\texttt{set.mm}\xspace}

\title{HyperTree Proof Search for Neural Theorem Proving}

\author{%
Guillaume Lample\thanks{Equal contribution. Corresponding authors: \nolinkurl{{glample,malachaux,tlacroix}@fb.com}}~ \thanks{Meta AI Research \quad $^{\ddagger}$Vrije Universiteit Amsterdam \quad $^{\mathsection}$ CERMICS \'Ecole des Ponts ParisTech}\\
\And
Marie-Anne Lachaux\textnormal{$^{*\dagger}$}\\
\And
Thibaut Lavril\textnormal{$^{*\dagger}$}\\
\And
Xavier Martinet\textnormal{$^{*\dagger}$}\\
\And
Amaury Hayat\textnormal{$^{\mathsection}$}\\
\And
Gabriel Ebner\textnormal{$^{\ddagger}$}\\
\And
Aur\'elien Rodriguez\textnormal{$^{\dagger}$}\\
\And
Timoth\'ee Lacroix\textnormal{$^{*\dagger}$}\\
}

\begin{document}
\maketitle

\newcommand{\inserteqsolvedidentities}{

\begin{table}[h!]
\centering
% \resizebox{\columnwidth}{!}{
\small
\begin{tabular}{l|cc|cc}
\toprule
         & \multicolumn{2}{c}{Proof size} & \multicolumn{2}{c}{Proof depth} \\
Identity & First & Best   & First & Best    \\

\midrule
$\exp(-x)\exp(x - y) = \exp(-y)$
& 6     & 6     & 6     & 6 \\
$\cosh(-x) = \cosh(x)$
& 4     & 4     & 4     & 4 \\
$\sin(\pi/2 + x) = \cos(x)$
& 8     & 8     & 8     & 7 \\
$0 < x \implies 2\ln(\sqrt{x}) = \ln(x)$
& 16    & 3     & 7     & 3 \\
$\cos(\pi/2 - x) = \sin(x)$
& 19    & 11    & 19    & 10 \\
$\sin(\pi/2 - x) = \cos(x)$
& 14    & 10    & 14    & 10 \\
$\cos(x)^2 + \sin(x)^2 = 1$
& 13    & 11    & 13    & 10 \\
$\cos(x) = \cos(x / 2)^2 - \sin(x / 2)^2$
& 16    & 11    & 16    & 7 \\
$\sin(x + y) - \sin(x - y) = 2\sin(y)\cos(x)$
& 24    & 14    & 23    & 14 \\
$0 < x \implies 2x \cosh(\ln(x)) = x^2 + 1$
& 20    & 14    & 18    & 12 \\
$\tanh(x) = (\exp(x) - \exp(-x)) / (\exp(x) + \exp(-x))$
& 46    & 23    & 30    & 11 \\
$\cos(x - y) + \cos(x + y) = 2\cos(x)\cos(y)$
& 33    & 19    & 33    & 13 \\
$\cosh(x) - \sinh(x) = \exp(-x)$
& 27    & 20    & 27    & 19 \\
$\cosh(x) - \sinh(x) = \frac{1}{\sinh(x) + \cosh(x)}$
& 55    & 38    & 40    & 20 \\
$\sin(2x) = 2\sin(x)\cos(x)$
& 27    & 15    & 19    & 8 \\
$\cos(2x) = 1 - 2\sin(x)^2$
& 130   & 27    & 118   & 21 \\
$\cosh(x - y) + \cosh(x + y) = 2\cosh(x)\cosh(y)$
& 84    & 31    & 84    & 29 \\
$\tanh(x) = (\exp(2x) - 1) / (\exp(2x) + 1)$
& 205   & 65    & 176   & 39 \\
$\sin(x) = 2\sin(x / 2)\cos(x / 2)$
& 29    & 17    & 21    & 8 \\
$\cos(2x) = 2\cos(x)^2 - 1$
& 72    & 26    & 68    & 19 \\
$\cos(x)^2 = 1 + \cos(2x) / 2$
& 71    & 30    & 61    & 16 \\
$\sinh(x) = 2\sinh(x / 2)\cosh(x / 2)$
& 64    & 37    & 51    & 25 \\
$\sinh(2x) = 2\sinh(x)\cosh(x)$
& 71    & 34    & 61    & 24 \\
$\sinh(x + y) = \sinh(x)\cosh(y) + \cosh(x)\sinh(y)$
& 130   & 77    & 121   & 63 \\
$\cosh(x - y) = \cosh(x)\cosh(y) - \sinh(x)\sinh(y)$
& 90    & 66    & 75    & 56 \\
$\cos(x + y)\cos(x - y) = \cos(x)^2 - \sin(y)^2$
& 117   & 64    & 117   & 64 \\
$\sin(x + y)\sin(y - x) = \cos(x)^2 - \cos(y)^2$
& 118   & 64    & 118   & 63 \\
$\lvert \sinh(x / 2)\rvert  = \sqrt{(\cosh(x) - 1) / 2}$
& 86    & 53    & 61    & 36 \\
$\sin(x + y)\sin(x - y) = \sin(x)^2 - \sin(y)^2$
& 183   & 66    & 183   & 65 \\
$\cosh(x)^2 = 1 + \cosh(2x) / 2$
& 87    & 40    & 71    & 32 \\
$\cosh(2x) = 2\cosh(x)^2 - 1$
& 78    & 42    & 62    & 33 \\
$\cosh(2x) = \cosh(x)^2 + \sinh(x)^2$
& 97    & 72    & 80    & 64 \\
$\tanh(x) - \tanh(y) = \sinh(x - y) / (\cosh(x)\cosh(y))$
& 154   & 135   & 85    & 81 \\
$\tanh(x) + \tanh(y) = \sinh(x + y) / (\cosh(x)\cosh(y))$
& 162   & 144   & 95    & 91 \\
$\sqrt{1 + \sinh(x)^2} = \cosh(x)$
& 82    & 70    & 76    & 62 \\
$\sin(x)^3 = (3\sin(x) - \sin(3x)) / 4$
& 72    & 58    & 63    & 49 \\
$\sin(3x) = 3\sin(x) - 4\sin(x)^3$
& 80    & 56    & 71    & 47 \\
$\cosh(3x) = 4 \cosh(x)^3 - 3\cosh(x)$
& 204   & 105   & 176   & 79 \\
$\cosh(x)^3 = (3\cosh(x) + \cosh(3x)) / 4$
& 162   & 106   & 137   & 79 \\
$\sin(4x) = \cos(x)(4\sin(x) - 8\sin(x)^3)$
& 73    & 73    & 60    & 60 \\
$\cos(\pi + x) = -\cos(x)$
& 148   & 28    & 118   & 9 \\
$\sin(\pi - x) = \sin(x)$
& 73    & 28    & 45    & 11 \\
$\cos(\pi / 3) = \sin(\pi / 6)$
& 26    & 17    & 26    & 17 \\
$\cos(\pi / 4) = \sin(\pi / 4)$
& 24    & 17    & 24    & 17 \\
$\cos(\pi / 6) = \sin(\pi / 3)$
& 22    & 17    & 22    & 17 \\
$\cos(2\pi + x) = \cos(x)$
& 125   & 70    & 37    & 18 \\
$\sin(2\pi + x) = \sin(x)$
& 353   & 69    & 62    & 16 \\
\bottomrule
\end{tabular}

\label{tab:solved_identities} % referenced
\vspace{0.3cm}
\caption{\textbf{Examples of identities solved.} Some of the 144 identities found by our model, in the order they were first solved. For each identity, we provide the size and the depth, both the for first proof, and for the minimal proof (i.e. the proof with the smaller number of steps) found during online training. The model found proofs with over 350 steps, some exceeding a depth of 100. After additional proof search, the model is often able to find shorter proofs. The proof of $\sin(2\pi + x) = \sin(x)$ requires a large number of steps, as the model can only use simple rules (e.g. the trigonometric rules provided in Table~\ref{tab:rules_trigo_full}), and it does not have access to the value of $\sin(2\pi)$ or $\sin(\pi)$.}
\end{table}

}

\begin{abstract}
We propose an online training procedure for a transformer-based automated theorem prover. Our approach leverages a new search algorithm, HyperTree Proof Search (\algo), inspired by the recent success of AlphaZero.
Our model learns from previous proof searches through online training, allowing it to generalize to domains far from the training distribution.
We report detailed ablations of our pipeline's main components by studying performance on three environments of increasing complexity.
In particular, we show that with \algo alone, a model trained on annotated proofs manages to prove $65.4\%$ of a held-out set of \mm theorems, significantly outperforming the previous state of the art of $56.5\%$~by~GPT-f.
Online training on these unproved theorems increases accuracy to $82.6\%$.
With a similar computational budget, we improve the state of the art on the \lean-based \minicurr dataset from $31\%$ to $42\%$ proving accuracy.
\end{abstract}

\section{Introduction}

\looseness=-1 Over the course of history, the complexity of mathematical proofs has increased dramatically.
The nineteenth century saw the emergence of proofs so involved that they could only be verified by a handful of specialists.
This limited peer review process inevitably led to invalid proofs, with mistakes sometimes remaining undiscovered for years (e.g. the erroneous proof of the Four Colour Conjecture~\cite{KempeOnTG}).
Some mathematicians argue that the frontier of mathematics has reached such a level of complexity that the traditional review process is no longer sufficient, envisioning a future where research articles are submitted with formal proofs so that the correctness can be delegated to a computer~\cite{voevodsky2011univalent}.

Unfortunately, very few mathematicians have adopted formal systems in their work, and as of today, only a fraction of existing mathematics has been formalized.
Several obstacles have hindered the widespread adoption of formal systems.
First, formalized mathematics are quite dissimilar from traditional mathematics, rather closer to source code written in a programming language, which makes formal systems difficult to use, especially for newcomers.
Second, formalizing an existing proof still involves significant effort and expertise (the formalization of the Kepler conjecture took over 20 person years to complete \cite{keplerhales}) and even seemingly simple statements sometimes remain frustratingly challenging to formalize.

\begin{figure}[h]
\centering
\vspace{0.1cm}
\includegraphics[width=1.0\textwidth]{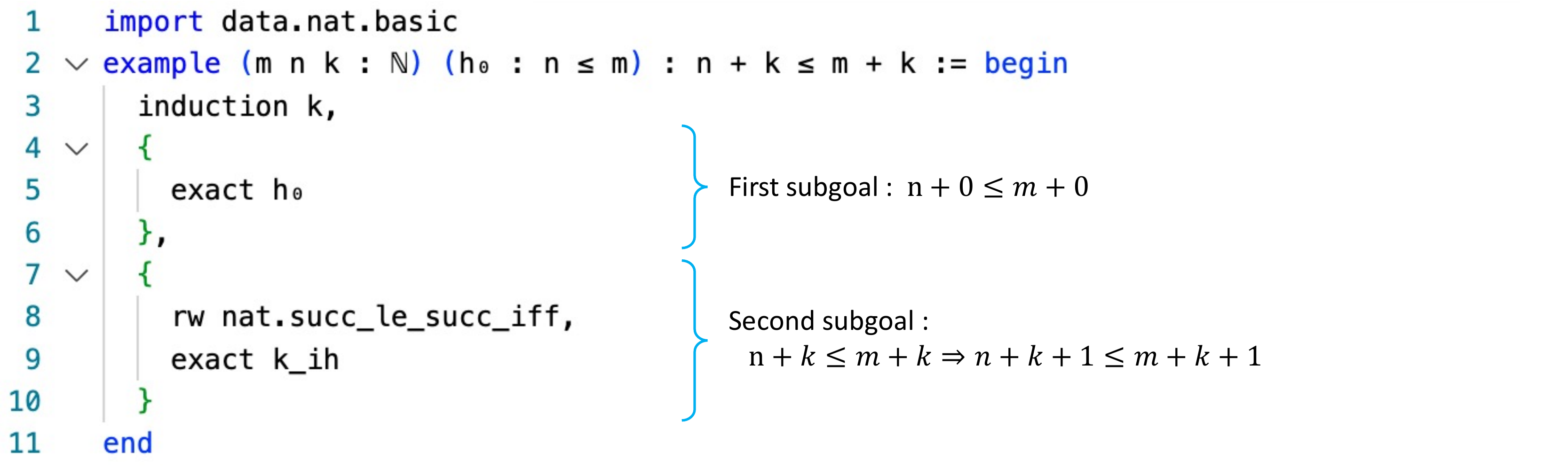}
\vspace{-0.2cm}
\caption{
\small
\textbf{A simple proof of the statement $\mathbf{n \leq m \Rightarrow n + k \leq m + k}$ in Lean.} The \textit{induction} tactic reduces the initial statement to two subgoals, that can be solved independently.
\label{fig:example_lean} % referenced
\vspace{0.2cm}
}
\end{figure}

To write a formal proof, mathematicians typically work with Interactive Theorem Provers (ITPs).
The most popular ITPs provide high-level ``tactics'' that can be applied on an input theorem (e.g. the initial goal) to generate a set of subgoals, with the guarantee that proving all subgoals will result in a proof of the initial goal (reaching an empty set means the tactic solves the goal).
An example of a proof in \lean~\cite{moura2015lean}, an interactive theorem prover, is given in Figure~\ref{fig:example_lean} and the corresponding proof hypertree is illustrated in Figure~\ref{fig:lean_proof_tree}. A tactic, \texttt{induction k}, is applied on the root goal ($n + k \leq m + k$) to start a proof by induction
\footnote{\looseness=-1 A hypergraph is a graph where an edge leads to a set of nodes that is potentially empty in our set-up. A hypertree is a hypergraph without cycles. More formal definitions can be found in Appendix~\ref{app:hypergraph_and_def}}.
The formal system returns two subgoals: $n + 0 \leq m + 0$ (the initial case) and $n + k \leq m + k \Rightarrow n + k + 1 \leq m + k + 1$ (the induction step).
As the first subgoal is our initial hypothesis, it can be solved using the \texttt{exact} tactic.
To prove the second subgoal, we first rewrite it using \texttt{nat.succ\_le\_succ\_iff}, a theorem from the \lean library stating that $m + 1\leq n + 1 \iff m \leq n$.
The new subgoal then becomes our induction hypothesis $n + k \leq m + k$, and can again be solved using the \texttt{exact} tactic, thereby solving the last remaining subgoal and completing the proof of the initial statement.
Starting from a goal and reducing it to subgoals until these can be solved, is commonly referred to as backward proving.

In this paper, we aim at creating a prover that can automatically solve input theorems by generating a sequence of suitable tactics without human interaction. Such a prover could significantly reduce the effort required to formalize existing mathematics.
The backward procedure naturally suggests a simple approach where a machine learning model trained to map goals to tactics interacts with an ITP to build the proof of an input goal in a backward fashion.
The automated prover builds a hypergraph with the theorem to be proved as the root node, tactics as edges and subgoals as nodes. The prover recursively expands leaves by generating tactics with our model until we find a proof of the initial theorem. A proof in this setup is a hypertree rooted in the initial theorem whose leaves are empty sets.
As many different tactics can be applied to a goal, and each tactic application can result in multiple subgoals, the number of goals in the graph grows exponentially and it is critical to reduce the search to the most promising branches. This can be done through techniques like alpha-beta pruning \cite{knuth1975analysis} or Monte Carlo Tree Search (MCTS)~\cite{abramson1987model}, known for its recent successes in difficult two player games~\cite{silver2018general}.
However, challenges arise in search algorithms for theorem proving that do not occur in two player games. For instance:
\vspace{0.1cm}

\begin{itemize}
    \item The action space, i.e. the amount of possible ``moves'' in a given state, is infinite (there is an unlimited number of tactics that can be applied to a given theorem). This requires sampling possible actions from a language model for which training data is scarce. Moreover, if all tactics sampled at a goal fail, we have no information on what region of the probability space to sample next.
    \item In the context of theorem proving, we need to provide a proof of all subgoals created by a tactic, whereas AlphaZero for two player games is allowed to focus on the most likely adversary moves.
    \item In Chess or Go, playing a sub-optimal move does not necessarily lead to losing the game, thus exploring these branches can provide information. In theorem proving, it is frequent to generate tactics that result in subgoals that can no longer be proved and on which the model will waste significant search budget.
\end{itemize}

This paper presents an in-depth study of our approach to overcome these difficulties and the resulting model, \model. In particular, we make the following contributions:

\begin{itemize}
    \item A new MCTS-inspired search algorithm for finding proofs in unbalanced hypergraphs.
    \item A new environment (\eq) to easily prototype and understand the behavior of the models we train and our proof search.
    \item A detailed ablation study and analysis of the different components used in our approach on three different theorem proving environments. We study how data is selected for training the policy model after a successful or failed proof-search, what target should be used to train the critic model, and the impact of online training vs. expert iteration.
    \item State-of-the-art performance on all analyzed environments. In particular, our model manages to prove over $82.6\%$ of proofs in a held-out set of theorems from \setmm in \mm, as well as $58.6\%$ on \minivalid~\cite{zheng2021minif2f} in \lean.
\end{itemize}

We begin by introducing related work in Section~\ref{sec:related_work} and present the three theorem proving environments that we study in Section~\ref{sec:proving_envs}. Then, we present our proof-search algorithm in Section~\ref{sec:proof_search}, our online training pipeline in Section~\ref{sec:online_training}, and all experimental details in Section~\ref{sec:experiments}. Finally, we describe our main results and ablation studies in Section~\ref{sec:results} before concluding in Section~\ref{sec:conclusion}.

\section{Related work}
\label{sec:related_work} % referenced

Automated theorem proving has been a long-standing goal of artificial intelligence, with the earliest work dating from the 1950s~\cite{gilmoreproofmethod, davisputnam}.
Early approaches focused on simpler logics, culminating in extremely efficient first-order provers such as E~\citep{schulz-e-2002} or Vampire~\citep{Riazanov2001Vampire1}.
However, these approaches are insufficient when it comes to theorems written in modern proof assistants such as Isabelle~\cite{nipkow-et-al-2002}, Coq~\cite{bertot2013interactive}, or \lean~\cite{moura2015lean}.
Recently, the rising success of deep language models~\cite{radford2019language} and model-guided search methods~\cite{silver2018general} has spurred a renewed interest in the problem of automated theorem proving.

\paragraph{Neural theorem provers.} Recent work applying deep learning methods to theorem proving \cite{polu2020generative,han2021proof,polu2022formal} are the closest to this work and obtained impressive results on difficult held-out sets for \mm and \lean. 
The main differences between their approach and ours are the proof-search algorithm we propose, the training data we extract from proof-searches and our use of online training compared to their expert iterations. We validate experimentally that these differences lead to improved performances as well as faster training times.
Another similar approach is Holophrasm~\cite{whalen2016holophrasm}, which is based on a different tree exploration technique which expands paths in an AND/OR tree, while we expand entire proof subtrees in a proof hypergraph. Their model is only trained once from supervised data and does not benefit from online training or expert iteration, which we found to be critical.
DeepHOL~\cite{bansal2019holist} focuses on the HOL-Light environment~\cite{harrison1996hol}. Their model relies on a classifier that can select among a restricted set of tactics and arguments, while we rely on a seq2seq model that can generate arbitrary tactics.
The suggested tactics are then used in a breadth-first search.
TacticToe~\cite{gauthier2021tactictoe} uses an MCTS without learned components, using ranking on predefined features to guide the search. Machine learning has also been used to improve classical provers by re-ranking clauses~\cite{chvalovsky2021learning}.
Overall, previous studies always focus on a single proving environment (e.g. \mm, \lean, or HOL-Light).
% In this paper, we extensively study the performance of our prover on three different formal languages, and found that some conclusions significantly vary based on the considered environment.

\paragraph{Reasoning abilities of language models.} Impressive performance of large language models in one or few shot learning \cite{radford2019language}, machine translation \cite{sutskever2014sequence} or more recently code generation \cite{lachaux2020unsupervised} spurred interest into the reasoning capabilities of large transformers. These model perform quite well on formal tasks such as expression simplification~\cite{saxton2018analysing}, solving differential equations~\cite{Lample2020Deep}, symbolic regression~\cite{d2022deep, petersen2021deep}, or predicting complex properties of mathematical objects~\cite{charton2020learning}. These studies suggest that deep neural networks are well adapted to complex tasks, especially when coupled with a formal system for verification.

\paragraph{MCTS and two player games.} \looseness=-1 Recently, AlphaZero~\cite{silver2018general} demonstrated good performances on two player games, replacing the Monte-Carlo evaluations of MCTS~\cite{abramson1987model} with evaluations from a deep neural network and guiding the search with an additional deep policy. This recent success follows extensive literature into search methods for two player games, notably alpha-beta search~\cite{knuth1975analysis}. Theorem proving can be thought of as computing game-theoretic value for positions in a min/max tree: for a goal to be proven, we need one move (max) that leads to subgoals that are all proven (min). Noticing heterogeneity in the arities of min or max nodes, we propose a search method that goes down simultaneously in all children of min nodes, such that every simulation could potentially result in a full proof-tree.

\section{Proving environments}
\label{sec:proving_envs} % referenced

In this paper, we develop and test our methods on three theorem proving environments: a) \mm, b) \lean and c) \eq. \mm~\cite{metamath} comes with a database of $30k$ human-written theorems called \setmm. We also evaluate our methods on the \lean proving environment, which provides a level of automation that is helpful to solve more complex theorems. \lean comes with a human-written library of $27k$ theorems called Mathlib~\cite{The_mathlib_Community_2020}. Finally, since \mm proofs can be quite difficult to understand and \lean requires more computing resources, we developed our own environment called \eq, restricted to proofs of mathematical identities. Its simplicity makes it ideal for prototyping and debugging. We briefly introduce these environments below.

\subsection{\mm}
\label{sec:metamath}

\begin{figure}[!h]
    \centering
    \includegraphics[width=0.7\textwidth]{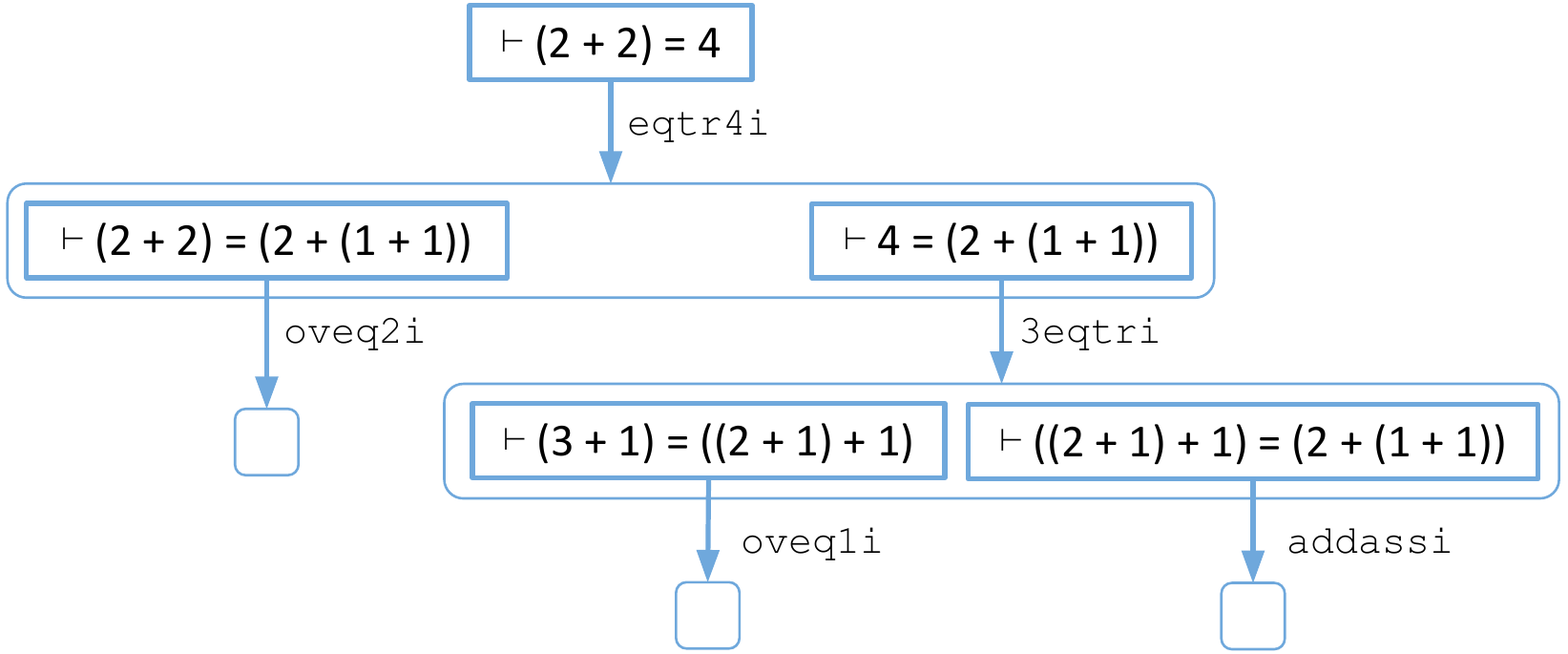}
    \caption{\small \textbf{A visualization of the proof-tree for $\mathbf{2+2=4}$ in \mm.}}
    \label{fig:mm_2p2e4} % referenced
\end{figure}

\mm's only rule is string substitution. Starting from a theorem to be proven, variables are substituted until we reach axioms. In our setup, we consider a tactic to be the label of a theorem in \setmm, along with the necessary substitutions. As shown in Figure~\ref{fig:mm_2p2e4}, to show that $2+2=4$, the first step uses \texttt{eqtr4i} which states that $A=B \wedge C=B \Rightarrow A=C$ with substitutions: $A = ( 2 + 2)$, $B = (2 + (1 + 1))$, and $C = 4$. We are then left with two subgoals to prove: $(2 + 2) = (2 + (1 + 1))$ and $4 = (2 + (1 + 1))$.

The simplicity of \mm makes it a great test bed for our algorithms. However, its lack of automation leads to larger proof sizes and its syntax and naming conventions make each step difficult to interpret for neophytes. Similar to GPT-f, we implement a parser for \mm in order to automatically prove the syntactic correctness of statements. Moreover, we use this parser to allow generating only substitutions that cannot be inferred from the goal.

\subsection{Lean}
\label{sec:lean}
\begin{figure}[h!]
    \centering
    \includegraphics[width=0.7\textwidth]{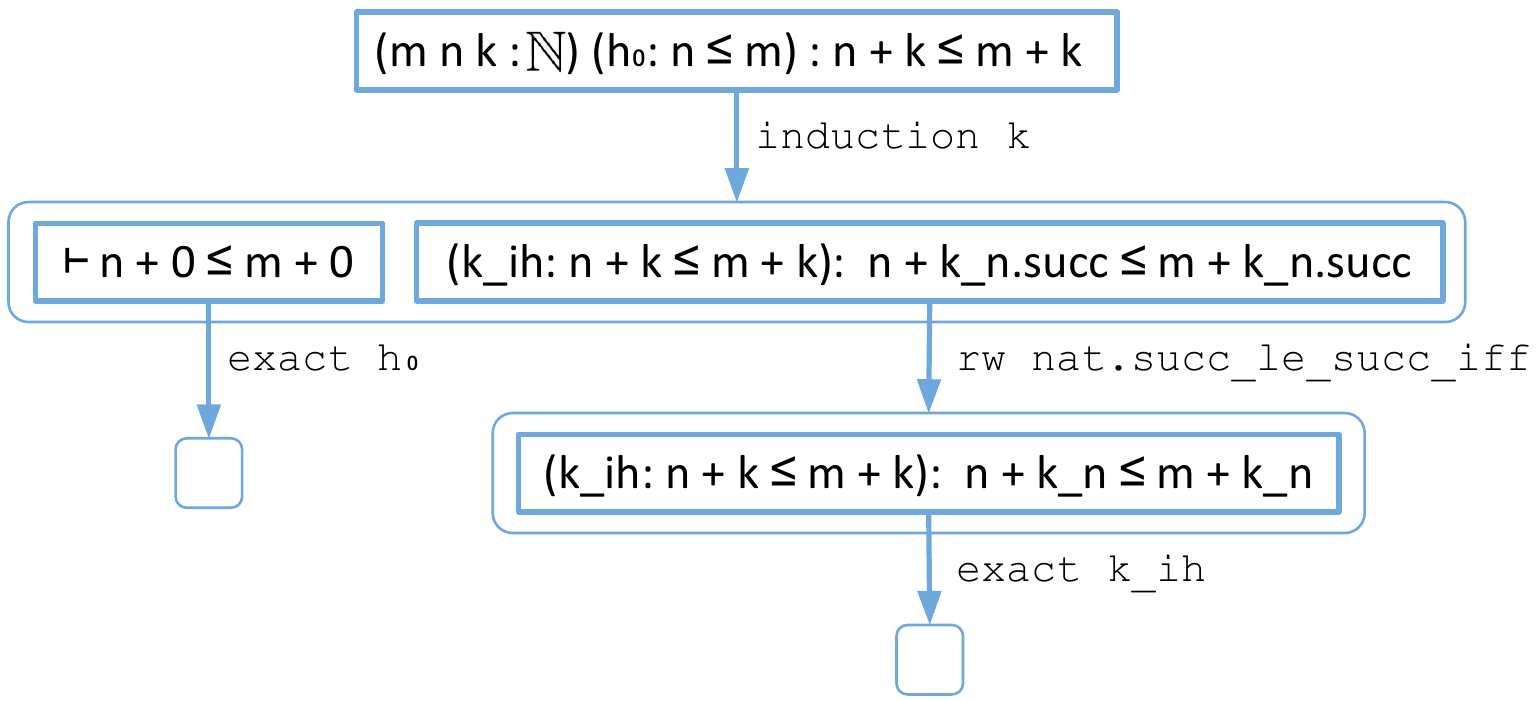}
    \caption{\small \textbf{A visualization of the proof-tree for the proof discussed in the introduction in \lean.}}
    \label{fig:lean_proof_tree} % referenced
\end{figure}

Lean is a full-fledged programming language and benefits from more powerful automation than \mm, with tactics such as \texttt{ring} (able to prove goals using manipulations in semirings), \texttt{norm\_num} (able to prove numerical goals) or \texttt{linarith} (able to find contradictions in a set of inequalities). An example of a Lean proof-tree is shown in Figure~\ref{fig:lean_proof_tree}.

States are more complex in \lean than in \mm: metavariables can appear which are holes in the proof to be filled later.
Subgoals sharing a metavariable cannot be solved in isolation. This is addressed in \citet{polu2020generative} by using as input the entire tactic state. Instead, we inspect tactic states to detect dependencies between subgoals, and split the tactic state into different subgoals where possible in order to maximize state re-use and parallelization in the proof search algorithm.

Finally, \lean's kernel type checker has to be called after each tactic application as tactics sometimes generate incorrect proofs and rely on the kernel for correctness. For every goal in the previous tactic state, we type check the proof term inserted by the tactic. Since the kernel does not support metavariables, we replace every metavariable by a lambda abstraction.

\subsection{Equations}
\label{sec:equations}
We developed the \eq environment as a simpler analogue to existing proving environments. Its expressivity is restricted to manipulating mathematical expressions (e.g. equalities or inequalities) with simple rules (e.g. $A+B=B+A$, or $A<B \Rightarrow -B<-A$). This reduced expressivity makes goals and tactics easy to understand, helping with interpretability and debugging: plotting the set of goals explored during a \mm proof search does not give a lot of insights on whether it is on track to find a proof.
In Section~\ref{appendix:eq} of the appendix, we give an in-depth presentation of this environment, of how we represent goals (Section~\ref{appendix:eq_th}), tactics (Section~\ref{appendix:eq_tac}) and how we prove statements (Section~\ref{appendix:eq_prove}).

Unlike in \mm or \lean, we do not have access to a training set of human annotated proofs for this environment.
Instead, we create a training set composed of randomly generated synthetic theorems and their proofs~(see Sections~\ref{appendix:eq_gen}~and~\ref{appendix:eq_gen} for details), and manually create an out-of-domain set of non-trivial mathematical identities for which we do not provide proofs, e.g. $\cosh(3x) = 4 \cosh(x)^3 - 3 \cosh(x)$ or $\sin(4x) = (4 \sin(x) - 8 \sin(x)^3)\cos(x)$. We refer to this evaluation split as \identities, a set of 160 mathematical expressions.

\begin{figure}[h!]
    \centering
    \includegraphics[width=0.5\textwidth]{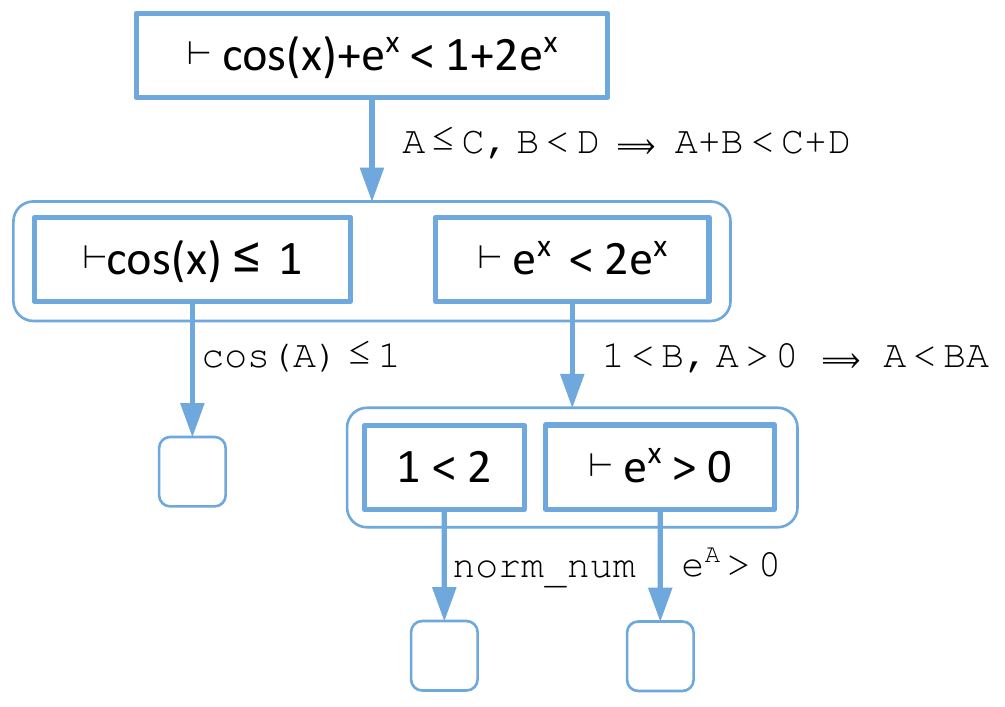}
    \caption{\small \textbf{A visualization of the proof-tree for $\mathbf{\cos(x)+e^x<1+2e^{x}}$ in \eq.}}
    \label{fig:proof_eq_basic} % referenced
\end{figure}

As synthetic theorems randomly generated are much simpler and significantly differ from statements in the \identities split, we can evaluate the ability of our model to generalize to complex and out of domains data. An example proof-tree in \eq is shown in Figure~\ref{fig:proof_eq_basic}.

\section{HyperTree Proof Search}
\label{sec:proof_search} % referenced in intro

\begin{figure*}[ht]
    \hspace{-0.8cm}
    \includegraphics[width=1.1\textwidth]{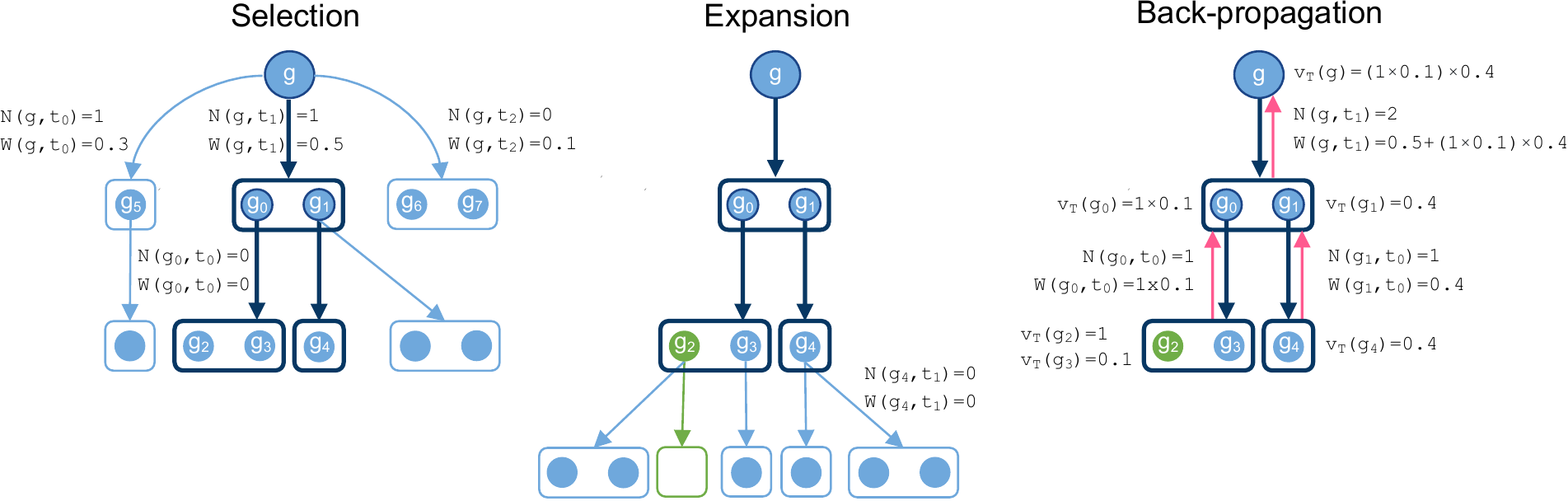}
    \vspace{-0.3cm}
    \caption{\small \textbf{HyperTree Proof Search}. We aim at finding a proof of the root theorem $g$ with \algo. Proving either $\{g_5\}$, $\{g_0,g_1\}$, or $\{g_6,g_7\}$ would lead to a proof of $g$ by tactic $t_0, t_1$, or $t_2$. The figure represents the three steps of \algo that are repeated until a proof is found. Guided by the search policy, we select a hypertree whose leaves are unexpanded nodes. The selected nodes are then expanded, adding new tactics and nodes to the hypergraph. Finally, during back-propagation we evaluate the node values of the hypertree, starting from the leaves back to the root, and update the visit counts and total action values.}
    \label{fig:main_fig} % referenced
\end{figure*}

Given a main goal $g$ to automatically prove, proof search is the algorithm that interacts with our learned model and the theorem proving environment to find a proof hypertree for $g$. Proof search progressively grows a hypergraph starting from $g$. A proof is found when there exists a hypertree from the root to leaves that are empty sets.

In this section, we assume a policy model $P_\theta$ and critic model $c_\theta$. Conditioned on a goal, the policy model allows the sampling of tactics, whereas the critic model estimates our ability to find a proof for this goal.
Our proof search algorithm will be guided by these two models. Additionally, and similar to MCTS, we store the visit count $N(g,t)$ (the number of times the tactic $t$ has been selected at node $g$) and the total action value $W(g,t)$ for each tactic $t$ of a goal $g$. These statistics will be used in the selection phase and accumulated during the back-propagation phase of the proof search described in Section~\ref{sec:selection} and Section~\ref{sec:backup}.

The algorithm iteratively repeats the three steps described below to grow the hypergraph until either a proof is found or we exceed our expansion budget. We show an example of these three steps of proof search in Figure~\ref{fig:main_fig}. We refer to this algorithm as  HyperTree Proof Search (\algo) throughout this work.
A more detailed comparison between \algo, MCTS, and the best-first search algorithm of \citet{polu2020generative} can be found in Appendix~\ref{sec:comparison_alg}.

\subsection{Selection}
\label{sec:selection} % referenced in section intro

The number of nodes in the proof hypergraph grows exponentially with the distance to the root. Thus, naive breadth-first search is infeasible to find deep proofs and some prioritization criteria is required to balance depth and breadth. This is the family of best-first search algorithms, of which \textit{A*} and \textit{MCTS} are instances.
Similar to \textit{MCTS}, we balance the policy model's prior with current estimates from the critic. In particular, we experiment with two different search policies: PUCT~\cite{silver2017mastering} and Regularized Policy (RP)~\cite{grill2020monte}, detailed in Appendix~\ref{appendix:policies}.
These search policies use the tactic prior from the policy model, the visit count $N$, and the estimated value of the tactic given by $Q=W/N$. A higher visit count will lead to a higher confidence in the estimated value than in the prior policy model, and \textit{vice-versa} for low visit counts.

The key difference between previous work and ours is that our proof search operates on a hypergraph.
Thus, whereas an algorithm like \textit{MCTS} will go down a path from the root to an unexpanded node during its selection phase, our algorithm will instead create a partial proof hypertree, leading to a set of either solved or unexpanded nodes.
The selection phase algorithm, described in more details in Appendix~\ref{appendix:algo_mcps}, consists in recursively following the search policy from the root until we find leaves of the current hypergraph.

In Figure~\ref{fig:main_fig}, we illustrate the selection step. We start at the root node $g$, which has three tactics $t_0, t_1, t_2$. The search policy selects $t_2$, leading to the set of subgoals $\{g_0,g_1\}$.
Then, for both $g_0$ and $g_1$, we again select the best tactic according to the search policy and finally reach the sets of unexpanded subgoals $\{g_2, g_3\}$ and $\{g_4\}$.
The final selected proof hypertree $T$ is composed of $g, \{g_0,g_1\}, \{g_2,g_3\}, \{g_4\}$ and is colored in dark blue in Figure~\ref{fig:main_fig}.

In order to batch calls to the policy and critic models over more nodes to expand, we run several selections sequentially, using a virtual loss~\cite{chaslot2008parallel, silver2018general} to produce different partial proof-trees.
Note that solving all unexpanded leaves of any of these trees would immediately lead to a full proof of the root. In the next section, we describe how nodes are expanded.

\subsection{Expansion}
\label{sec:expansion}
To expand a node $g$, we use the policy model to suggest tactics that would make progress on the goal, then evaluate these tactic suggestions in the theorem proving environment. Each valid tactic will lead to a set of new subgoals to solve, or to an empty set if the tactic solves the goal. Finally, we add a hyperedge for each valid tactic $t_i$ from the expanded node $g$ to its (potentially empty) set of children for this tactic $\{g_i^0,...,g_i^k\}$. Note that these children might already be part of the hypergraph. For new nodes, visit counts $N(g,t)$ and total action values $W(g,t)$ are initialized to zero.
There are three types of nodes in the hypergraph:
\begin{itemize}
    \item \textit{Solved}: at least one tactic leads to an empty set, or has all its children solved.
    \item \textit{Invalid}: all tactics sampled from the policy model were rejected by the environment, or lead to invalid nodes.
    \item \textit{Unsolved}: neither solved nor invalid, some tactics have unexpanded descendants.
\end{itemize}
Note that the definitions for \textit{Solved} or \textit{Invalid} are recursive. These status are updated throughout the hypergraph anytime a hyperedge is added. Tactics leading to invalid nodes are removed to prevent simulations from reaching infeasible nodes. Once this is done, we back-propagate values from the expanded nodes up to the root, as described in the next section.

In Figure~\ref{fig:main_fig}, we show an example of expansion. After selecting a hypertree $T$ during the selection step, for each unexpanded leaf goal of $T$, we generate $B=2$ tactics with our policy model and keep only the valid ones. This results in two tactics for $g_2$ and $g_4$ and one for $g_3$. We apply these tactics in our formal environment and obtain new sets of subgoals and add them to the hypergraph. One tactic of $g_2$ solves $g_2$, resulting in an empty set of subgoals. Note that because $g_3$ is not solved yet, $g_0$ remains unsolved.

\subsection{Back-propagation}
\label{sec:backup} % referenced in section intro

For each expanded goal $g$ in a simulated proof tree $T$, its value is set to $v_{T}(g) = 1$ if it is solved, and $v_{T}(g) = 0$ if it is invalid. Otherwise, its value is estimated by the critic model: $v_T(g)=c_\theta(g)$. This provides $v_T$ for all leaves of $T$ and we can then back-propagate in topographic order (children before parents) through all nodes of $T$.
Interpreting the value of a node as the probability that it can be solved, since solving a goal requires solving all of its children subgoals, the value of a parent is the product of values of its children (we assume that the solvability of subgoals is independent, for simplicity):
$$v_T(g) = \prod_{c \in \textrm{children}(g,t)}v_T(c)$$ 

In particular, the value of a goal $g$ is the product of the values of all leaves in $T$ that remain to be solved to obtain a proof of $g$. Once all values in $T$ are computed, we increment the corresponding visit count $N(g,t)$ in the hypergraph as well as the total action values: $W(g,t) \pluseq v_T(g)$. For a goal $g$, the estimated value for tactic $t$ is then the mean of the total action value:

$$Q(g,t) = \frac{W(g,t)}{N(g,t)}$$

We give an example of back-propagation in Figure~\ref{fig:main_fig}. First, we evaluate the values of the leaf nodes of $T$. Because $g_2$ is solved, we set $v_T(g_2)=1$. The values of $g_3$ and $g_4$ are estimated with the critic model, e.g. $v_T(g_3)=c_\theta(g_3)=0.1$. The values of the internal nodes are obtained by computing the product of their children values. Thus, we first compute $v_T(g_0) = v_T(g_2) \times v_T(g_3)$ and $v_T(g_1) = v_T(g_4)$, then $v_T(g)=v_T(g_0) \times v_T(g_1) = (v_T(g_2) \times v_T(g_3)) \times (v_T(g_4))$. Then, for every (goal, tactic) pair $(g, t)$ in $T$, we increment the visit count, $N(g,t) \pluseq 1$ and update the total action value:  $W(g,t) \pluseq v_T(g)$.

\section{Online training from proof searches}
\label{sec:online_training} % referenced

In the previous section, we considered the policy and critic models as given. In this section, we explain how proof search is used to create training data for these two models. Provers are asynchronously running proof searches using a version of the models synchronized with the trainers, coupling training and data extraction in an online procedure that leads to continuous performance improvements.

\subsection{Training objectives}
\label{sec:training_obj}

Both the policy model $P_\theta$ and the critic model $c_\theta$ are encoder-decoder transformers~\cite{vaswani2017attention} with shared weights $\theta$, which are trained with a tactic objective and a critic objective respectively.

\paragraph{Tactic objective.}
The policy model $P_\theta$ takes as input a tokenized goal and generates tactics. It is trained with a standard seq2seq objective~\cite{sutskever2014sequence}, where we minimize the cross-entropy loss of predicted tactic tokens conditioned on an input goal.

\paragraph{Critic objective.}
In order to decode a floating point value with our seq2seq critic model $c_\theta$, we start decoding with a special token, restrict the output vocabulary to the two tokens $\texttt{PROVABLE}$ and $\texttt{UNPROVABLE}$, and evaluate the critic with $c_\theta(g) = P(\texttt{PROVABLE} | g, \texttt{CRITIC})$.
The critic objective is identical to a seq2seq objective where the cross-entropy is minimized over the two special tokens.

\subsection{Online training}
\label{sec:online_training_online_training}
We use a distributed learning architecture reminiscent of AlphaZero~\cite{silver2017mastering} or distributed reinforcement learning setups~\cite{nair2015massively, silver2018general}. A distributed data parallel trainer receives training data from a set of asynchronous provers that run proof searches on tasks chosen by a controller that also centralizes statistics. Provers, in turn, continuously retrieve the latest model versions produced by the trainers in order to improve the quality of their proof search. This set-up is represented in Figure~\ref{fig:online_arch}.
Once a prover finishes a proof-search, we extract two types of training samples from its hypergraph:

\begin{figure}
\centering
\includegraphics[width=0.5\textwidth]{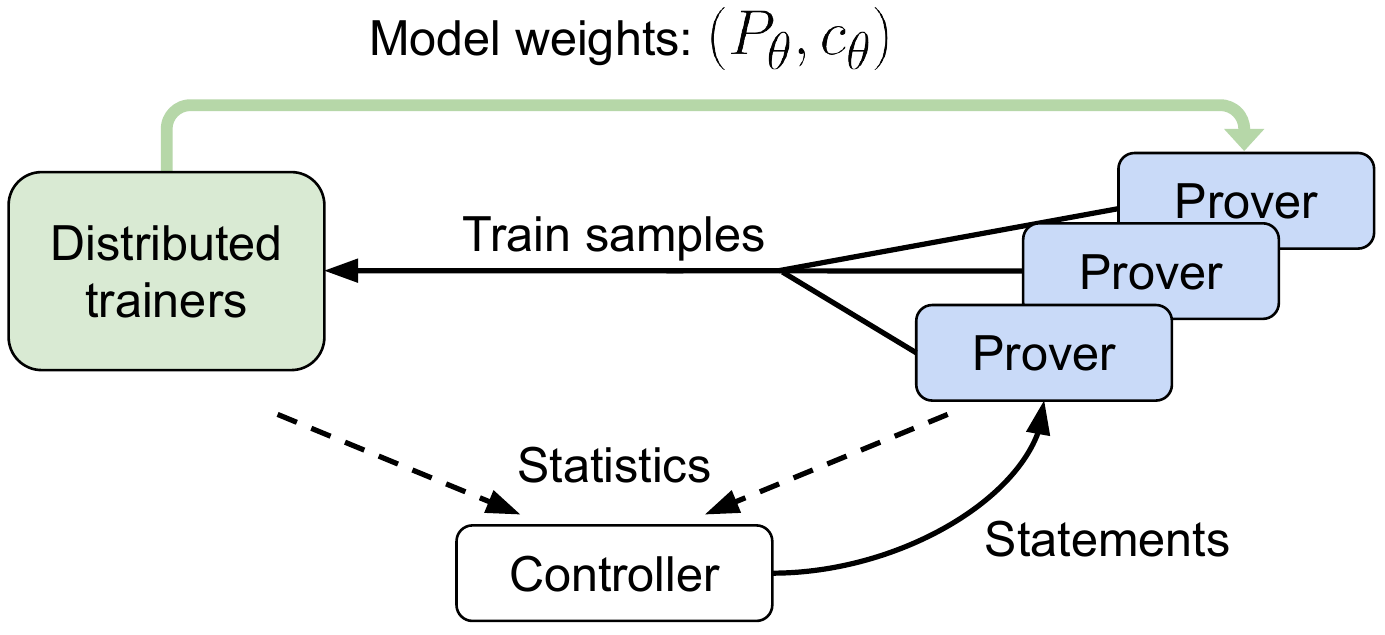}
\caption{\small \textbf{An overview of our online training architecture.} The controller sends statements to asynchronous \algo provers and gathers training and proving statistics. The provers send training samples to the distributed trainers and periodically synchronize their copy of the models.}
\label{fig:online_arch} % referenced
\vspace{-0.2cm}
\end{figure}

\paragraph{Tactic samples.} \looseness=-1 At the end of a successful proof search, we extract (goal, tactic) pairs of a minimal proof hypertree of the root node as training samples for the policy model. This selection has a large impact on performances, other options such as selecting all solved nodes are investigated in Section~\ref{sec:ablationdata}. We use a different minimality criteria depending on the environment: total number of proof steps for \mm and \eq, and total tactic CPU time for \lean (see Appendix~\ref{appendix:time_vs_depth} for details).

\paragraph{Critic samples.} In the proof search hypergraph, we select all nodes that are either solved, invalid, or with a visit count higher than a threshold.
Then, we use $c(g)=1$ as the training target for solved nodes. For internal nodes, we use the final estimated action value $c(g)=W(g,t^*)/N(g,t^*)$ where $t^*$ is the tactic that maximizes the search policy at $g$. Finally, for invalid nodes, we use $c(g)=0$.

The trainers receive training samples that are stored into two separate finite-size queues, one for each objective.
When a queue is full, appending a new sample discards the oldest one.
In order to create a batch for a task, we uniformly select samples in the corresponding queue.
The two training objectives are weighted equally.
Additionally, during online training, we continue sampling from the supervised tasks which provide high-quality data.

Our proof-search depends on many hyper-parameters, and the optimal settings might not be the same for all statements, making tuning impractical. Thus, the controller samples these hyper-parameters from pre-defined ranges (see Appendix~\ref{appendix:hyperopt} for details) for each different proof-search attempt.

\subsection{Full training pipeline}
\label{sec:full_training_pipeline}

In order to bootstrap our online learning procedure we require a policy model $P_\theta$ that outputs coherent tactics. While the critic is left untrained, the policy model is fine-tuned on a pretrained transformer using a supervised dataset specific to the target environment. Overall, the full training pipeline can be summarized as follows:

\begin{itemize}
    \item \textbf{Pretraining} of the encoder-decoder model on a large unsupervised corpus (c.f. Section~\ref{sec:modelpretraining}).
    \item \textbf{Fine-tuning} of the policy model on supervised datasets detailed in (c.f. Section~\ref{sec:finetuning}).
    \item \textbf{Online training} of both the policy and critic models on data extracted from proof search.
\end{itemize}

\section{Experiments}
\label{sec:experiments} % referenced

In this section, we provide details about our experimental training and evaluation protocols. We first describe the supervised datasets used to fine-tune our policy models, as well as the tokenization used. We then give practical details on pretraining and the model architecture. Finally, we discuss the evaluation datasets and methodology.

\subsection{Model fine-tuning and supervised datasets}
\label{sec:finetuning} % referenced

Starting the \algo procedure described in Section~\ref{sec:online_training} from a randomly initialized model would be sub-optimal, as no valid tactic would ever be sampled from the policy model. Thus, starting the online training from a non-trivial model is critical. To this end, we first fine-tune our policy model $P_\theta$ on a supervised dataset of theorems specific to each environment.

\paragraph{\mm} In \mm, we extract all proofs from the \setmm library, composed of 37091 theorems (c.f. Section~\ref{appendix:hash_versions} for the version of \setmm). We first derive a graph of dependencies between statements, and generate a random train-valid-test split of theorems, with 1000 valid and test theorems. We use the DAG to ensure that each theorem in the valid or test set is not used to prove another theorem.
Moreover, this DAG is used to build a table of \textit{forbidden tokens}: if the proof of $A$ depends on $B$, we set to zero the probability of generating the token $A$ during a proof-search of $B$.
We use a seq2seq training objective, where the model is conditioned on a goal to prove, and is trained to output a sequence of the following format:
$$\texttt{LABEL MANDATORY\_SUBSTS <EOU> LABEL\_STATEMENT PREDICTABLE\_SUBSTS <EOS>}$$
\texttt{LABEL} is the label of the rule to apply, \texttt{MANDATORY\_SUBSTS} is a serialized version of the substitutions in the rule that cannot be inferred from syntactic parsing of the input goal and the theorem statement.
During proof-search, decoding is stopped at the \texttt{<EOU>} (End Of Useful) token and we do not generate predictable substitutions, as this would unnecessarily increase decoding time and the probability that our model generates invalid substitutions.
Training the model to output predictable substitutions and the rule statement serves as a co-training task and helps reduce overfitting. The training set is composed of around 1M goal-tactic pairs; more statistics about the training data are provided in Table~\ref{tab:supervised_stats}. Tokenization in \mm is trivial, as statements are composed of space-separated tokens.

\begin{table*} % [!h]
    \begin{center}
    \begin{tabular}{lccc}
    \toprule
           & \# train theorems & \# train proof steps & Avg. goal length \\
    \midrule
    \eq    &  $\infty$         & $\infty$             & 33.7             \\
    \mm    &  35k              & 1M                   & 120.1            \\
    \lean  &  24k              & 144k                 & 169.3            \\
    \bottomrule
    \end{tabular}
    \label{tab:supervised_stats} % referenced
    \caption{\small \textbf{Dataset statistics for supervised training.}}
    \end{center}
    % \vspace{-0.3cm}
\end{table*}

\paragraph{\lean} Following \cite{polu2022formal}, we extract a supervised dataset from the Mathlib library. The training set is composed of 24k theorems and 144k goal-tactic pairs. In addition, we co-train with the dataset of proof-artifacts of \citet{han2021proof} to reduce overfitting.
To facilitate experimentation and reproducibility, we use fixed versions of \lean, Mathlib, and miniF2F (c.f. Appendix~\ref{appendix:hash_versions}).
Finally, we add another supervised co-training task by converting to \lean a synthetic dataset of theorems generated by the \eq environment (c.f. Appendix~\ref{sec:eq_to_lean}).
In order to avoid hooking into the \lean parser, we tokenize goals and tactics using byte-pair encoding (BPE~\cite{sennrich2015neural}) following previous work~\citep{polu2020generative, polu2022formal}. Statistics about the training set are available in Table~\ref{tab:supervised_stats}.

\paragraph{\eq} Unlike \mm or \lean, the \eq environment does not come with with a dataset of manually annotated proofs of theorems.
Instead, we generate supervised data on the fly using the random graph generator described in Appendix~\ref{appendix:eq_gen}.
As the model quickly reaches a 100\% proving accuracy on these synthetic theorems, there would be no benefit in using them during online training.
Thus, we fine-tune on the synthetic dataset, and only leverage statements from the \identities split during online training.
As in \mm, tokenization of statements for this environment is natural, as each statement can be tokenized using the list of symbols from its prefix decomposition~\cite{Lample2020Deep}.

\subsection{Model pretraining}
\label{sec:modelpretraining}

Model pretraining can be critical in low-resource scenarios where the amount of supervised data is limited~\cite{devlin2018bert, lample2019cross}. Thus, we do not immediately fine-tune our model but first pretrain it on a large dataset to reduce overfitting and improve generalization.
In particular, we pretrain our model with a masked seq2seq objective (MASS~\cite{song2019mass}) on the LaTeX source code of papers from the mathematical section of arXiv.
After tokenization, our filtered arXiv dataset contains around 6 billion tokens for 40GB of data.
Similar to \citet{polu2020generative}, we observed large performance gains using pretraining.
However, we found that arXiv alone provides a better pretraining than when it is combined with other sources of data (e.g. GitHub, Math StackExchange, or CommonCrawl).

\subsection{Model Architecture and Training}
\label{sec:architecture_and_training}

\paragraph{Model architecture.} Our transformer architecture uses a 12-layer encoder and a 6-layer decoder in all experiments. We use an embedding dimension of 1600 in the encoder and 1024 in the decoder for both \mm and \lean. For \eq, where we expect the model to require less decoding capacity, the decoding dimension is lowered to 512.
We found that reducing the decoder capacity increases the decoding speed without impacting the performance, as previously observed by~\citet{kasai2020deep} in the context of machine translation.
Our models are composed of 440M parameters for \eq and 600M parameters for \mm and \lean (for comparison, GPT-f uses a 770M parameter, 36-layer model).

\paragraph{Supervised fine-tuning.} During fine-tuning, we train our models with the Adam optimizer~\cite{kingma2014adam} and an inverse square-root learning rate scheduler~\cite{vaswani2017attention}.
We use a dropout of 0.2~\cite{srivastava2014dropout} to reduce the overfitting of our models.
We also apply layer-dropout \cite{layerdrop} with a dropout rate of 0.1 to further reduce overfitting and stabilize training.
We implement our models in PyTorch \cite{paszke2017automatic} and use float16 operations to speed up training and to reduce the memory usage of our models.

\paragraph{Online training.} During online training, we alternate between the \textit{goal-tactic} objective, used during fine-tuning on the supervised dataset, and the \textit{goal-tactic} and \textit{goal-critic} objectives on data generated by the provers.
As the model and the data generated by the provers are constantly evolving, we do not want the learning rate to decrease to 0, and we fix it to $3 \times 10^{-5}$ after the warm-up phase.
Unless mentioned otherwise (e.g. for large experiments), we run all \mm and \eq experiments with 16 trainers and 32 provers for a total of 48 V100 GPUs.

\subsection{Evaluation settings and protocol}
\label{sec:eval_settings_protocol}

In \citet{polu2022formal}, the model is fine-tuned on theorems from the training set and expert iteration is done on theorems from different sources: train theorems, synthetic statements, and an extra curriculum of statements without proofs (\minicurr).
The produced model is then evaluated on unseen statements, namely the validation and test splits of the miniF2F dataset~\cite{zheng2021minif2f}.

In this work, we also consider the \textit{transductive} setup: on a corpus of unproved statements available at train time, how many proofs can our method learn to generate?
This protocol is also sensible, as allowing the model to learn from a failed proof-search can lead to more focused exploration on the next attempt, proving more statements overall than a model that would not be trained online.

Following~\cite{polu2020generative}, we also evaluate the pass@k by running $k$ proof searches on the evaluated statements with the policy and critic obtained by online training.
In the transductive setup, we also report the \textit{cumulative pass rate}, i.e. the proportion of theorems solved at least once during online training.

\section{Results}
\label{sec:results} % referenced
In this section, we present our results and study the moving parts of our pipeline through ablations.
Each experiment is run on a single environment (e.g. \lean, \mm, or \eq). 
We compare our model with GPT-f\cite{polu2020generative,han2021proof,polu2022formal} which represents the state of the art on \mm and \lean.

\subsection{Main Results}
\label{sec:main_results}

\begin{table*}[!h]
\begin{center}
\small
\begin{tabular}{l|c|ccc|c}
\toprule
                            & Supervised & GPT-f & \model-1d & \model-7d & \model\\
    Online training statements & -  &  \multicolumn{3}{c|}{\minicurr} & \minivalid  \\
    \midrule
    \minivalid  & 38.5 & 47.3 & 46.7  & 47.5 & 58.6$^\dagger$ \\
    \minitest   & 35.3 & 36.6 & 38.9  & 40.6 & 41.0 \\
    \minicurr   & 20.8 & 30.6 & 33.6$^\dagger$ & 42.5$^\dagger$ &  32.1 \\
    \midrule
    Train time (A100 days)     & 50 & 2000 & 230 & 1620 &  1360 \\
    \bottomrule
\end{tabular}

\end{center}
\caption{
\small
\textbf{Pass rate on \lean environment using 64 trials (pass@64).} Numbers with a $^\dagger$ exponent correspond to the cumulative pass-rate since the evaluated statements are part of the online training. \model refers to the method described in this paper.
}
\end{table*}

\subsubsection{Lean}
\label{sec:lean_results}

In \lean, we run our experiments on A100 GPUs with 32 trainers and 200 provers. Each prover runs our \lean API on 48 CPU cores.
Unlike \citet{polu2022formal}, we sample statements equally from \mathlibtrain and \minicurr, to avoid giving too much importance to statements from a different domain than the target.
After 1 day of training (i.e. $(200 + 32)$ A100 days of compute), each statement from \minicurr has been sampled on average $250$ times, and $110$ out of the $327$ statements have been solved. Our model outperforms GPT-f on \minitest, with an approximately $10\times$ training time speed-up. After $7$ days, we solve $139$ statements of \minicurr ($100$ for GPT-f), and observe further improvements on \minivalid or \minitest.

For other evaluations, we depart from the set-up of~\citet{polu2022formal}, directly using the statements from the \minivalid split in our online training, obtaining 58.6\% cumulative pass rate. We then evaluate the final model on \minitest, reaching 41\% pass@64, against 36.6\% for GPT-f.

Without the synthetic data co-training task, the performance drops to $54.9\%$ cumulative pass rate on the \minivalid split, and $38.5\%$ pass@64 on the \minitest split. Examples of proofs found by our model can be found in Appendix~\ref{app:lean_proofs}.

\subsubsection{\mm}

On \mm, we train our model on V100 GPUs, with 128 trainers and 256 provers, whereas ablations are run on 16 trainers and 32 provers.
We report our results in Table~\ref{tab:mm_main_results} for the supervised model and for a model trained with online training.
During online training, we sample equally statements from the training and from the validation splits of \setmm.

\begin{figure}
\centering
\includegraphics[width=1.0\textwidth]{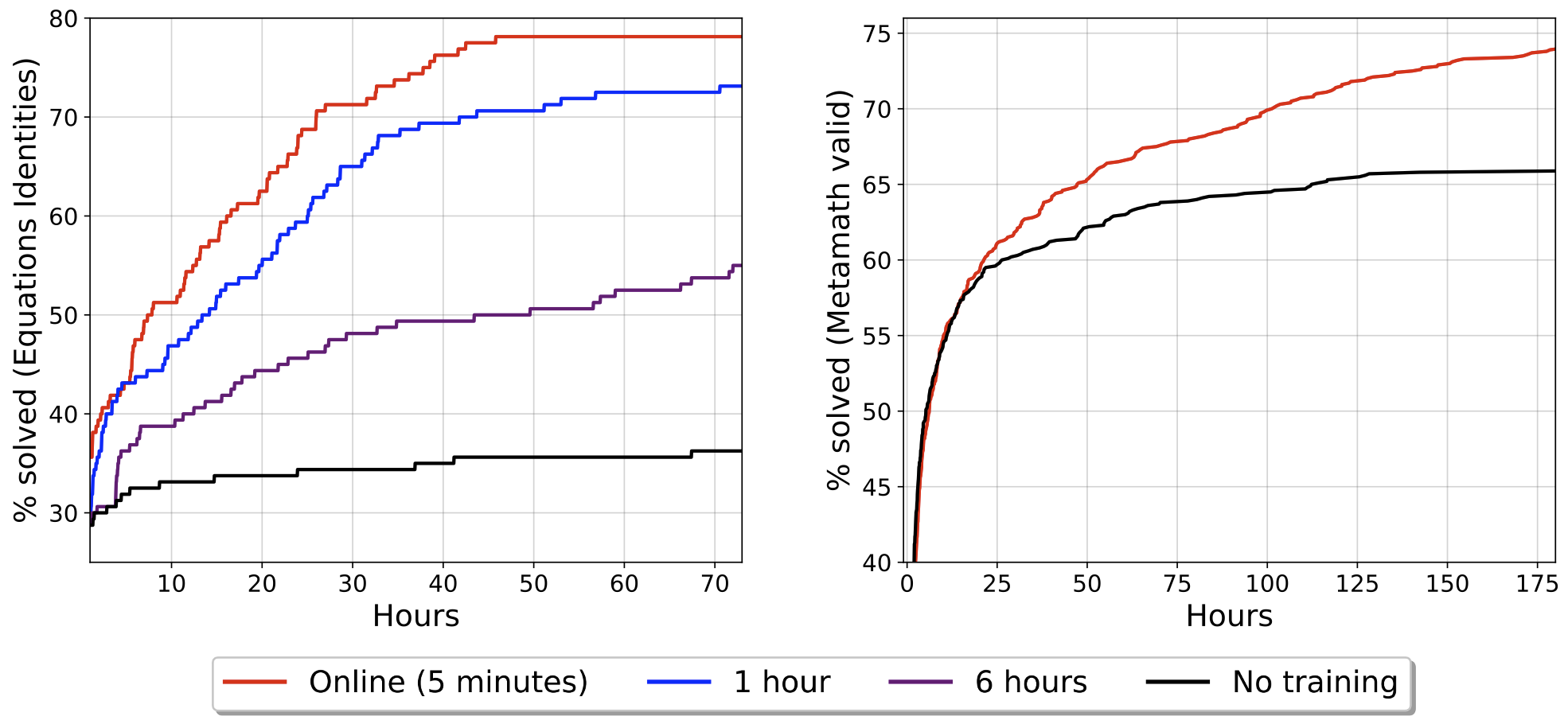}
\caption{
\small
\looseness=-1 \textbf{Comparison between online setup, expert iteration, and fixed model.} We report the cumulative pass rate on the \identities (resp. valid) split for the \eq (resp. \mm) environment.
Reloading the model more frequently converges faster and to a better performance. When ``No training'' is done (i.e. the model is the supervised one), the final performance is much lower despite using as many attempts. This shows that online training is able to learn from previous proof searches.}
\label{fig:eq_mm_ablation_reload_freq} % referenced
\end{figure}

\looseness=-1 Online training dramatically improves performances on valid statements, going from a $61\%$ pass@8 to a cumulative pass rate of $82.6\%$ on this split.
This improvement cannot solely be explained by the high number of attempts on validation theorems during training. Indeed, the ablation in Figure~\ref{fig:eq_mm_ablation_reload_freq} (right) shows that \model significantly outperforms a supervised model with the same number of attempts. The supervised model plateaus at $66\%$ while \model keeps improving beyond $74\%$ after 7 days of training, showing that the model is able to learn from previous proof searches through online training.

\looseness=-1 On test theorems, for which statements were not provided during online training, the pass@32 accuracy increased by $10\%$ compared to the supervised model, from $55.8\%$ to $65.6\%$. Note that the supervised model already obtains an accuracy of $65.4\%$ (resp. $61.2\%$) on the validation (resp. test) split, compared to GPT-f's $56.5\%$ (resp. $56.2\%$) after expert iteration, showing the benefits of \algo.

\begin{table*}[!h]
\begin{center}
% \resizebox{0.7\columnwidth}{!}{
\small
\begin{tabular}{l|ccc|cc}
\toprule
      &   \multicolumn{3}{c|}{Valid}   & \multicolumn{2}{c}{Test} \\
      &  cumulative & pass@8 & pass@32 & pass@8 & pass@32 \\
    \midrule
    % Supervised (3k) & N/A    & 58.6\% & 62.2\% & 54.4\% & 58.6\% \\
    Supervised      & N/A    & 61.0\% & 65.4\% & 55.8\% & 61.2\% \\
    % \model     (3k) & 82.6\% & 81.0\% & 81.0\% & 64.4\% & 67.8\% \\
    \model          & 82.6\% & 81.0\% & 81.2\% & 65.6\% & 72.4\% \\
    \bottomrule
\end{tabular}
% }
\end{center}
\caption{
\small
\textbf{Results on \mm for a supervised model and \model.}
We report the pass@8 and pass@32 scores on the validation and test splits.
Additionally, for \model we also report the cumulative score on the validation set, i.e. the fraction of theorems proved at least one time during online training. Note that for \model on Valid, the cumulative and pass@k performances are close since these statements were seen during training.
}
\label{tab:mm_main_results} % referenced
\end{table*}

\subsubsection{\eq}

In \eq, we run our main experiment with 32 trainers and 64 provers, whereas ablations are run on 16 trainers and 32 provers. In this environment, the model easily learns the training distribution of our random generator, and solves all synthetically generated problems. Thus, online training is run on the Identities statements only. Our main experiment reaches a cumulative pass rate of $91.3\%$ on the \identities split, while a supervised model never exceeds $36\%$ even after a similar number of proof attempts. In Appendix~\ref{tab:solved_identities}, we give examples of \identities statements proved during online training, as well as the size and depth of proofs found by the model.

In particular, \model managed to find the proof of complex mathematical statements, such as \smash{$\sinh(x / 2) = \sinh(x) / \sqrt{2(1+\cosh(x))}$} and $\tan(3  x)(1-3(\tan(x))^2) = 3  \tan(x)-(\tan(x))^2  \tan(x)$ that required 82 and 117 proof steps respectively, showing the abilities of \algo to prioritize subgoals and guide the search in very large proof graphs.
This shows that online training is able to adapt our policy and critic models to a completely new domain, going from automatically generated statements to identities found in math books.
Examples to understand the gap between these two domains can be found in Appendix~\ref{appendix:eq}.

\subsection{Ablation study}
\label{sec:ablations}

In this section, we present an ablation study on several components of our system.
Since \lean experiments are CPU intensive, we run most of our ablations on the \eq and \mm environments.
On \lean, we ran experiments on a smaller subset of hyper-parameters that consistently performed well on the other environments.

\subsubsection{Online training data for tactic objective}
\label{sec:ablationdata} % referenced

\begin{table*}[!h]
\begin{center}
\small
\begin{tabular}{l|cccc|c}
\toprule
    Proof Of         & \multicolumn{2}{c}{All Solved} & \multicolumn{2}{c|}{Root} & All Nodes \\
    Type of Proof    & All  & Min                     & All  & Min           & - \\
    \midrule
    \mm (valid)      & 61.2 & 65                      & 57.4 & \textbf{68.6} & 51.6 \\
    \mm (test)       & 57.2  & \textbf{58.8}          & 54.8 & 57.4          & 54.4 \\
    \midrule
    \eq (Identities) & 40.6 & \textbf{78.1}           & 37.5 & 71.3          & 37.5 \\
    \bottomrule
\end{tabular}
\end{center}
\caption{
\small
\textbf{Performance of our model for different online training data for tactic objective} We report the pass@8 score for \mm and cumulative pass rate for \eq. We try to keep all nodes and sample the tactics of these nodes according to the policy. We also try to extract proofs or minimal proofs of solved nodes, or the proofs or minimal proofs of the root theorem only. Selecting minimal proofs always improves performances and gives the best results in both environments.}
\label{tab:ablation_data} % referenced
\end{table*}

The way we filter tactics sent to the trainers has a large impact on final performances.
We investigated several filtering methods and report the results in Table~\ref{tab:ablation_data}.
The first method is similar to the one used in AlphaZero and exposed in \cite{silver2017mastering}: we select all nodes of the proof search hypergraph where the visit count is above a certain threshold and we filter tactics above a given search policy score. At training time, tactics are sampled according to the filtered search policy. With this method the model reaches 51.6\% pass@8 on the valid set of \mm and 37.5\% cumulative pass rate on Equations. 

We then experimented with other filtering criteria, selecting only goal-tactic pairs that are part of proofs: either a proof of the root node, or of any solved node in the hypergraph. Then, we learn from all possible proofs, or only from proofs that are minimal according to a criteria (number of proof steps for \eq and \mm, cumulative CPU time for \lean).

\looseness=-1 Learning only from the minimal proofs always leads to improved performance, regardless of the selected roots.
Learning from the minimal proofs of all solved nodes, we reach a cumulative pass rate of  78.1\% on \eq, compared to 40.6\% when learning from all proofs.
On \mm, only learning from the root's minimal proof gives the best result on the validation set, reaching a pass@8 of 68.6\%.

\subsubsection{Critic}
\label{sec:ablationcritic} % referenced

\begin{table*}[!h]
\begin{center}
\resizebox{0.8\columnwidth}{!}{
\small
\begin{tabular}{l|c|cc|c}
\toprule
    &  \model &  No critic & Hard critic   & Fixed search params \\
    \midrule
    \mm (valid)  & 68.6          & 64.8 & 67.6 & \textbf{69.8} \\
    \mm (test)   & \textbf{57.4} & 52.2 & 57.4  & 56.2 \\
    \midrule
    \eq (\identities) & \textbf{78.1} & 65.6 & 63.1  & 73.8 \\
    \bottomrule
\end{tabular}
}
\end{center}
\caption{
\small
\textbf{Ablation study on the critic and search hyper-parameters in \algo.}
We report the pass@8 score for \mm, and the cumulative pass rate for \eq.
\model, trained with a soft critic and stochastic hyper-parameters, obtains the best performance in both environments. Removing the critic, or using a hard critic (e.g. when the critic is trained to predict 1 on solved nodes and 0 on others) leads to reduced performances. In \eq, adding stochasticity in the proof search hyper-parameters increases the performance by $4.3\%$ in \eq, and slightly improves performance in \mm.
}
\label{tab:critichyperopt} % referenced
\end{table*}

\looseness=-1 To measure the impact of our critic model, we run an experiment where the proof search is only guided by the policy model.
During the back-propagation phase, we set $v_T(g)$ to $0.5$ for the leaves of $T$.
In that context, our model is no longer trained with a critic objective.
We run this experiment for both \eq and \mm, and report the results in Table~\ref{tab:critichyperopt}.
In both environments, using a critic model improved the performance significantly, by $5.2\%$ and $12.5\%$ on \mm and \eq respectively.

As mentioned in Section~\ref{sec:online_training}, to train the critic objective, we set the training targets as $c(g)=1$ for solved nodes, $c(g)=0$ for invalid nodes and $c(g)=W(g,t^*)/N(g,t^*)$ where $t^*$ is the tactic that maximizes the search policy at $g$., for internal nodes. We also tested a hard critic estimation of the target values, following \citet{polu2020generative}, where $c(g)=1$ for solved nodes and $c(g)=0$ for both invalid and internal nodes. We report results in Table~\ref{tab:critichyperopt}. For both \mm and \eq, estimating the critic target of internal nodes with the \algo action value estimate allows \model to reach its best performance. In \eq, the model reaches a cumulative pass rate of 78.1\%, compared to 63.1\% with hard critic estimates. In \eq, using hard critic targets gives worse performances than having no critic model at all, showing that these targets are a bad estimation: setting all internal nodes to zero is too pessimistic.

\subsubsection{Fixed proof search parameters}
\label{sec:ablation_model_freq}

We study the impact of sampling \algo hyper-parameters for each attempt during online training. We run experiments with fixed, chosen search parameters for \eq and \mm to compare with random sampling, and report results in Table~\ref{tab:critichyperopt}. \model achieves better performances than the model trained with fixed search parameters on \mm test set and \eq \identities, reaching 78.1\% pass rate compared to 73.8\% in \eq \identities.

\subsubsection{Model update frequency during online training}
\label{sec:ablation_model_reload}

In our online training procedure, the policy and critic models are updated every five minutes on the provers. We measure the impact of the frequency of these updates by trying different refresh rates: 5 minutes, 1 hour, 6 hours for \eq, and no updates at all for both \eq and \mm. We report the cumulative pass rate over training hours in Figure~\ref{fig:eq_mm_ablation_reload_freq}. The higher the refresh rate, the better the cumulative pass rate over time, confirming the benefits of online training over expert iteration.

\section{Conclusion}
\label{sec:conclusion} % referenced

\looseness=-1 In this work, we introduce \algo, an AlphaZero-inspired proof search algorithm for automated theorem proving, along with an online training procedure.
We run an extensive study of our pipeline, and present state-of-the-art results on multiple proving environments.
We show that online training provides large speed-ups over expert iteration, and allows generalization of the policy and critic models to completely new domains.
Despite large number of attempts per theorem, proving the entirety of datasets like \miniff remains elusive, and generating data from proof-search on the currently available corpora will likely be insufficient in the long term. 
As manually annotated formal datasets are limited, another way of providing exploration and additional training data (in the spirit of self-play for two player games) is required. Automated generation of new theorems is likely to be one of the future milestones.

\subsubsection*{Acknowledgments}
We thank the Meta AI and FLARE teams for useful comments and discussions throughout this work, notably, Faisal Azhar, Antoine Bordes, Quentin Carbonneaux, Maxime Darrin, Alexander Miller, Vincent Siles, Joe Spisak and Pierre-Yves Strub. 
We also thank the members of the Lean community for their help, notably Fabian Glöckle for valuable feedback on this project.

\bibliographystyle{unsrtnat}
\bibliography{biblio}

%%%%%%%%%%%%%%%%%%%%%%%%%%%%%%%%%%%%%%%%%%%%%%%%%%%%%%%%%%%%
\newpage

\appendix

%\section{Appendix}

\section{Proof search in more details}
\label{app:proof_search_details}

\subsection{Hypergraph and definitions}
\label{app:hypergraph_and_def}

We begin with some useful notations and concepts for our hypergraphs.

Formally, let $\mathcal{G}$ be a set of nodes, and $\mathcal{T}$ a set of tactics. A hypergraph is a tuple $H=(G, r, T, U)$ with $G\subset\mathcal{G}$ the nodes, $r\in G$ the root, and $T \subset G\times\mathcal{T}\times \mathcal{P}(G)$ the admissible tactics. An element of $T$ is written $(g, t, c)$ where $g$ is the start goal, $t$ is the applied tactic and $c$ is the potentially empty set of children that the tactic creates when applied to $g$ in the proving environment.

A hypertree is a hypergraph without cycles, i.e, such that we cannot find a path $g_0, \dots, g_\ell=g_0$ with $\ell >0$ and with $g_{i+1}$ in the children of $g_i$ for all $i$'s.

Let $S\subset G$ be the set of solved nodes. A node $g\in G\setminus U$ is solved if one of its tactic leads to no subgoals, or one of its tactics leads to only solved nodes. Formally: $\exists (g, t, \emptyset) \in T$ or $\exists (g, t, c) \in T$ such that $c \subset S$. We say that a tactic $t$ is \textit{solving} for $g$ if all the children it leads to are solved.
Conversely, let $U\subset G$ be the set of invalid nodes. A node $g\in G\setminus U$ is invalid if it has been expanded but has no tactics in the hypergraph, or all of its tactics have an invalid child. Formally: $\{(g, t, c) \in T\}=\emptyset$ or $\forall (g, t, c) \in T$, $c\cap I\neq\emptyset$.

These recursive definitions naturally lead to algorithms \texttt{MaintainSolved} and \texttt{MaintainStatus} to maintain sets $S$ and $I$ when elements are added to $H$.

A sub-hypertree $H_T$ of $H$ is a connected hypertree rooted at some goal of $H$. Its leaves $\textrm{leaves}(H_T)$ are its subgoals without children (either elements of $U$ or $S$). The set of \textit{proofs} of $g$ in $H$, $\textrm{Proofs}(g, H)$ are all the hypertrees rooted at $g$ that have all their leaves in $S$. Similarly, the \textit{expandable} subtrees of $H$ rooted in $g$, $\textrm{Expandable}(g,H)$ are the subtrees with at least one leaf in $U$. A tactic is said to be \textit{expandable} if it is part of an expandable subtree, this can be computed with a graph-search \texttt{ComputeExpandable}.

We can now reformulate the process of proof-search. Starting from a hypergraph that contains only the root theorem $r$, we produce a sequence of \textit{expandable} subtrees. The unexpanded leaves of these subtrees are expanded in the hypergraph, then the new value estimates are backed-up. The hypergraph grows until we use all our expansion budget, or we find a proof of $r$.

\subsection{Policies}
\label{appendix:policies} % referenced

When a goal $g$ is added to the hypergraph, its visit count $N(g,t)$ and total action value $W(g,t)$ are initialized to zero. Its virtual visit count $VC(g,t)$ are updated during proof search. Let $C(g,t) = N(g,t) + VC(g,t)$ be the total counts. These values are used to define the value estimate with a constant \textit{first play urgency} \cite{wang2007modifications}: 
\begin{equation*}
Q(g, t) = \left\{
    \begin{array}{ll}
        \frac{\max(1, N(g,t))}{\max(1, C(g,t))} & \mbox{if $t$ is solving for $g$} \\
        \frac{0.5}{\max(1,C(g,t))} & \mbox{if } N(g,t) = 0 \\
        \frac{W(g,t)}{C(g,t)} & \mbox{otherwise.}
    \end{array}
\right.    
\end{equation*}
Notice that the value of solving tactics decreases with virtual counts, allowing exploration of already solved subtrees.

Given the visit count $N$, the total counts $C$, value estimates $Q$, the model prior $P_\theta$ and an exploration constant $c$. The policy used in Alpha-Zero is $\textrm{PUCT}$:
\begin{equation*}
    \textrm{PUCT}(g) = \textrm{arg}\max_{t\in\mathcal{A}}\left [Q(g, t) + c \cdot P_\theta(t|g) \cdot \frac{\sqrt{\sum N(g,\cdot)}}{1 + C(g,t)}\right]
\end{equation*}
Notice that more weight is given to the value estimate $Q$ as $N$ grows which decreases the second term. Another work \cite{grill2020monte} obtains good performances using as search policy the greedy policy regularized by the prior policy.
\begin{equation*}
        \pi_{RP}(g) = \textrm{arg}\max_{y\in\mathcal{S}}\left[Q(g)^Ty - c\cdot\frac{\sqrt{\sum C(g,\cdot)}}{\sum (C(g,\cdot) + 1)} KL(\pi_\theta,y)\right]\quad{\textrm{with $\mathcal{S}$ the policy simplex at $g$}}
\end{equation*}
Again, note that this policy balances the prior with the value estimates as the count grows, but does not account for individual disparities of visits of each tactics. In our experiments, we obtained better performances with $\pi_{RP}$ on Equations, and better performances with $PUCT$ on \mm and \lean.

\subsection{Algorithms}
\label{appendix:algo_mcps} % referenced

\paragraph{Simulation} \looseness=-1 During simulation, we only consider subtrees that could become proofs once expanded. This means we cannot consider any invalid nodes or consider subgraphs containing cycles. If we encounter a tactic that creates a cycle during a simulation, this tactic is removed from the hypergraph, virtual counts from this simulation are removed and we restart the search from the root. This may remove some valid proofs, but does not require a backup through the entire partial subtree which would lead to underestimating the value of all ancestors. Removing tactics from the hypergraph also invalidates computations of \textit{expandable} tactics. This is dealt with by periodically calling \texttt{MaintainExpandable} if no valid simulation can be found. A full description of the algorithm that finds one expandable subtree is available in Algorithm~\ref{algo:one_simulation}.
Selection of nodes to expand requires finding expandable subtrees until a maximum number of simulations is reached, or no expandable tactic exists at the root.

\begin{algorithm}
\caption{Finding an expandable subtree}
\label{algo:one_simulation} % referenced
\begin{algorithmic}
\STATE {\bf Input:} A hypergraph $H$ and its $root$
\STATE {\bf Output:} A partial proof tree with unexpanded leaves
\STATE \textit{:start}
\STATE T: hypertree($root$)
\STATE to\_explore: list = [$root$]
\WHILE{to\_explore}
    \STATE g = to\_explore.pop()
    \IF{g is internal}
        \IF {$\textrm{expandable}(g) \neq \emptyset$}
            \STATE tactic = $\textrm{arg}\max_t \restr{\pi}{\textrm{expandable}(g)}\pi(g, t)$
        \ELSE
            \STATE continue \COMMENT{ expandable nodes are in a sibling branch }
        \ENDIF  
        \IF {tactic leads to cycle}
            \STATE kill tactic
            \STATE remove virtual counts for elements of T
            \STATE goto \textit{start}
        \ENDIF
        \STATE $VC(g, tactic) \pluseq 1$
        \STATE T.add(g, tactic, children(g, tactic))
        \STATE to\_explore += [children(g, tactic)]
    \ENDIF
\ENDWHILE
\end{algorithmic}
\end{algorithm}

\paragraph{Expansion} The policy model produces tactics for an unexpanded node $g$. These tactics are evaluated in the proving environments. Valid tactics are filtered to keep a unique tactic (e.g. the fastest in \lean) among those leading to the same set of children. Finally, we add the filtered tactics and their children to the hypergraph. If no tactics are valid, the node is marked as invalid and we call \texttt{MaintainInvalid}. If a tactic solves $g$, the node is marked as solved and we call \texttt{MaintainSolved}.

\paragraph{Backup}
The backup follows topological order from the leaves of a simulated partial proof-tree $T$, updates $W$ and $N$, and removes the added virtual count. The algorithm is described in Algorithm~\ref{algo:backup} % precisely
\begin{algorithm}
\caption{Back-propagation of total action value $W$}
\label{algo:backup} % referenced
\begin{algorithmic}
\STATE {\bf Input: } Partial proof-tree $T$ and value estimates $c_\theta(g)$ of its leaves.
\STATE to\_backup = []
\FOR{$g$ in leaves of T}
    \STATE $v_T(g) = c_\theta(g)$
    \STATE to\_backup.append($\textrm{parent}_T(g)$)
\ENDFOR
\WHILE{to\_backup}
\STATE $g$ = to\_backup.pop()
\STATE to\_update = $\prod_{c \in \textrm{children}_T(g)}$ $v_T(c)$
\STATE $W(g,t) \pluseq$ to\_update
\STATE $N(g,t) \pluseq 1$
\STATE $VC(g,t) \mineq 1$
\STATE $v(g,t) = $ to\_update
\STATE $g$.is\_prop = true
\IF{all $c$.is\_prop for $c$ in $\textrm{siblings}_T(g)$}
    \STATE to\_backup.append($\textrm{parent}_T(g)$)
\ENDIF
\ENDWHILE
\end{algorithmic}
\end{algorithm}

\subsection{Comparison with other search algorithms}
\label{sec:comparison_alg} % referenced

\paragraph{Best First Search~\cite{polu2020generative}.} This best-first search expands goals one at a time according to a priority-queue of either a value model or the cumulative log-prior from the language model.
Since the priority is equal among siblings but strictly decreasing with depth, this means siblings will always be expanded together.
However, nothing prevents the algorithm from jumping from one potential proof-tree to another, and potentially favoring breadth over depth.
In comparison, depth does not appear in the value estimate we compute, but rather the remaining number of nodes to solve a particular proof-tree. Moreover, our algorithm leads to value estimates that can be used to train our critic, which performs better than 0-1 estimates provided by best-first search (c.f. Section~\ref{sec:ablationcritic}).

\paragraph{Monte Carlo Tree Search~\cite{abramson1987model}.} \looseness=-1 MCTS has been famously used as part of AlphaZero~\cite{silver2017mastering} to obtain great performances on two player games. This two player set-up can be mapped to theorem-proving by assigning one player to choosing the best tactics while the other player picks the most difficult goal to solve (a method explored in Holophrasm~\cite{whalen2016holophrasm}).
However, since we need to provide a proof of the root theorem, we need to ensure that we can solve \textit{all} goals that a tactic leads to. This set-up has been studied for two player games when attempting to compute the game-theoretical value of positions.
Using MCTS in this set-up is suboptimal~\cite{winands2008monte}, ignoring unlikely but critical moves from the opponent (in our case, a subgoal that looks easy but is impossible to solve).
We decided to exploit the highly asymmetrical arities of our two players (most tactics lead to one or two goals) which makes simulating partial proof-trees computationally feasible.
Thus, the values we back-propagate always take into account all possible moves from the opponent, while only requiring a few expansions per simulation.

\section{\eq environment}
\label{appendix:eq} % referenced

In this section, we give additional details about the environment \eq. First, we described its main elements, theorems (resp. tactics) in Section~\ref{appendix:eq_th} (resp. \ref{appendix:eq_tac}). Then, we describe a proof in this environment in Section~\ref{appendix:eq_prove}, how numerical expressions are evaluated in Section~\ref{sec:eq_verif_numeric} and what vulnerabilities this can lead to in Section~\ref{sec:env_vulnerabilities}. Finally, we describe our random theorem generator in Section~\ref{appendix:eq_gen} and how theorems and their proofs can be translated to \lean in Section~\ref{sec:eq_to_lean}.

\subsection{Theorems}
\label{appendix:eq_th} % referenced

Each theorem in \eq consists in proving mathematical expressions composed of functions of real numbers, by manipulating and rewriting expressions.
A theorem to prove can be an inequality or an equality, conditioned to a set (potentially empty) of initial assumptions. For instance:

\begin{equation*}
    x^2 + 1 \geq 2x \;\;\;\; \textrm{or} \;\;\;\; x > y \implies e^{y - x} - 1 < 0
\end{equation*}

In the first example, the goal does not have any hypothesis and consists in proving that for every $x \in \mathbb{R}$, $x^2 + 1 \geq 2x$. In the second example, the goal consists in proving that $e^{y - x} - 1< 0$ for every $x, y \in \mathbb{R}$ that satisfy the hypothesis $x > y$.

Equalities and inequalities are represented as trees with the three following elements:

\begin{itemize}
\item \textbf{Leaves:} represent a variable, an integer, or a constant (e.g. $\pi$).
\item \textbf{Internal nodes:} represent unary or binary operators, e.g. $+$, $-$, $/$, $\times$, $\exp$, $\ln$, $\cos$, $\sin$, $\sinh$, $\cosh$, etc.
More advanced operators such as $\gcd$, $\mathrm{lcm}$, $\mathrm{mod}$ (the rest of an euclidean division) are possible when dealing with integers.
\item \textbf{A root node:} represents a comparison operator, e.g. $=$, $\leq$, $<$, $\geq$, $>$, $\neq$. More advanced comparison operators such as $|$ (divides) are possible when dealing with integers.
\end{itemize} 

\subsection{Tactics}
\label{appendix:eq_tac} % referenced 

\eq allows to deduce equalities and inequalities from simpler subgoals, using elementary rules (i.e. tactics).
The environment contains two types of rules: transformations, which consist in matching a pattern in an expression and replacing it by an equivalent expression; and assertions, which consist in asserting that an expression is true. Both types of rules can have assumptions.

\paragraph{Transformation rules}
A transformation rule (\texttt{TRule}) consists in a set of two expressions, $L$ and $R$, equivalent under a set of assumptions $S$. For instance $\texttt{TRule}(A+B, B+A)$ is the transformation rule stating the commutativity of the addition, namely that $A+B = B+A$ for any expressions $A$ and $B$. Note that in this case, the set of assumption $S$ is empty as the equality always holds. Another example is $\texttt{TRule}(\sqrt{A^2},A,[A \geq 0])$ that states that $\sqrt{A^2}=A$ provided that $A \geq 0$.

Applying such a rule to an existing equation works as follows:

\begin{itemize}
    \item matching a term $T$ in the expression that has the pattern of $L$
    \item identifying the matching variables and substituting them in $R$
    \item replacing $T$ by R in the input equation
    \item return the resulting equation with the set of hypotheses required for the transformation
\end{itemize}

For instance, if the input goal is:
$$\sqrt{({e^x})^2} = e^x$$

Applying $\texttt{TRule}(\sqrt{A^2},A,[A \geq 0])$ on this expression will result in two subgoals:
\begin{itemize}
    \item The same expression, where $\sqrt{A^2}$ has been replaced by $A$: $e^x = e^x$
    \item The hypothesis required for the assumption to hold: $e^x \geq 0$
\end{itemize}

More generally, a transformation rule will result in $N + 1$ subgoals, where $N$ is the number of hypotheses required by the rule.

\paragraph{Assertion rules}

An assertion rule (\texttt{ARule}) expresses the fact that an expression is true, provided some hypotheses. It is represented by a main expression, and a set of assumptions sufficient for the main expression to hold. For instance, the rule $\texttt{ARule}(A \leq C, [A \leq B, B \leq C])$
states the transitivity of the partial order $\leq$, i.e. $A\leq C$ provided that there exists an expression $B$ such that $A\leq B$ and $B\leq C$.

Assertion rules do not always have hypotheses, for instance the reflexivity rule $\texttt{ARule}(A = A)$, or the rule $\texttt{ARule}(e^A > 0)$ stating that $e^A$ is positive, for any real value $A$. Note that the two subgoals generated in the previous paragraph ($e^x = e^x$ and $e^x > 0$) can be respectively solved by these two assertion rules (i.e. by matching $A=e^x$ and $A=x$).

Unlike transformation rules that always result in at least one subgoal (the initial expression on which we applied the transformation), assertion rules will only generate $N$ subgoals, where $N$ is the number of hypotheses. As a result, being able to apply an assertion rule without hypotheses to an expression is enough to close (e.g. solve) the goal. Assertion rules are in fact very similar to rules in \mm.

In Table~\ref{tab:rules_per_type}, we provide the number of \eq rules in different categories.
Some examples of transformation and assertion rules are given in Table \ref{tab:rules_trigo_full}.

\begin{table}[h!]
\centering
\small
\newcolumntype{x}[1]{>{\centering\arraybackslash\hspace{0pt}}p{#1}}
\begin{tabular}{lccccc}
\toprule
 Rule type & Basic & Exponential & Trigonometry & Hyperbolic & All \\
 \midrule
 Transformation & 74 & 18 & 9 & 8 & 109 \\
 Assertion      & 90 & 11 & 9 & 0 & 110 \\
 \midrule
 Total & 171 & 29 & 18 & 11 & 219 \\
 \bottomrule
\end{tabular}
\vspace{0.3cm}
\caption{\textbf{Number of \eq rules in each category.}}
\label{tab:rules_per_type} % referenced
\end{table}

\begin{table}[h!]
\centering
\small
\begin{tabular}{ll}
\toprule
Transformation rules & Assertion rules \\
\midrule
$\sin(0) = 0$ & $ |\cos(A)| \leq 1$ \\
$\cos(0) = 1$ & $ |\sin(A)| \leq 1$ \\
$ \sin(\frac{\pi}{2}) = 1$ & $ |\sin(A)| \leq |A|$ \\
$ \cos(\frac{\pi}{2}) = 0$ & $A = B \implies \sin(A) = \sin(B)$ \\
$ \sin(-A) = -\sin(A)$     & $A = B \implies \cos(A) = \cos(B)$ \\
$ \cos(-A) = \cos(A)$                                      & $\sin(A) \neq \sin(B) \implies A \neq B$ \\
$\cos(A) \neq 0 \implies \tan(A) =\frac{\sin(A)}{\cos(A)}$ & $\cos(A) \neq \cos(B) \implies A \neq B$ \\
$\sin(A + B) = \sin(B) \cos(A) + \sin(A) \cos(B)$  & $A = B, \cos(A) \neq 0 \implies \tan(A) = \tan(B)$ \\
$ \cos(A + B) = \cos(A) \cos(B) - \sin(A) \sin(B)$ & $\tan(A) \neq \tan(B), \cos(A) \cos(B) \neq 0 \implies A \neq B$ \\
\bottomrule
\end{tabular}
\vspace{0.2cm}
\caption{\textbf{Trigonometric rules} accessible by the model. The model only has access to these elementary rules when proving statements from \identities. In particular, it cannot use more involved theorems such as $\cos^{2}(x)+\sin^{2}(x)=1$ or $\sin(\pi) = 0$.}
\label{tab:rules_trigo_full} % referenced in appendix
\end{table}

\subsection{Proving a statement with \eq}
\label{appendix:eq_prove} % referenced

In order to prove a theorem with \eq, the user (or automated prover) has to apply tactics on the current expression. A tactic can correspond either to a transformation rule, or to an assertion rule.

For transformation rules, the model needs to provide:
\begin{itemize}
    \item the rule (using a token identifier)
    \item the direction in which the rule is applied (a Boolean symbol, for forward or backward)
    \item an integer that represents the position where the rule is applied
    \item an optional list of variables to specify (c.f. paragraph below)
\end{itemize}

The direction of the rule indicates whether we want to transform $L$ by $R$ or $R$ by $L$ (e.g. replace $A$ by \smash{$\sqrt{A^2}$}, or the opposite).
The position where the rule is applied is given by the prefix decomposition of the input expression. For instance, the prefix notation of $(x + y) + 1$ is given by \texttt{+ + x y 1}.
Applying the commutativity rule $A+B=B+A$ to the expression in position 0 will result in $1 + (x + y)$. Applying it in position 1 will result in $(y + x) + 1$, since the rule was applied to $(x+y)$.
Note that for the commutativity rule, the direction in which we apply the rule does not matter.
The list of variables to specify is required when variables in the target patterns are absent from the source pattern. For instance, applying the transformation rule $\texttt{TRule(A,A+B-B)}$ in the forward direction will require to provide the value of $B$.

\looseness=-1 For assertion rules, the format is simpler. We no longer need to specify a direction or a position (the position is always 0 as the assertion statement must match the expression to prove), but only:
\begin{itemize}
    \item the rule (using a token identifier)
    \item an optional list of variables to specify
\end{itemize}

In this case, the list of variables to specify corresponds to variables that appear in hypotheses and that cannot be inferred from the main expression.
For instance, to apply the assertion rule $A \leq B \wedge B \leq C \implies A \leq C$, we need to specify the value of $B$.
We will then be left with two subgoals: $A \leq B$ and $B \leq C$.

Proving a statement in \eq requires to recursively apply tactics to unproved subgoals, until we are left with no subgoals to prove.
An example of proof-tree in \eq is shown in Figure~\ref{fig:proof_eq_basic}.
Figure~\ref{fig:simple_stuff_long} shows an example of proof of the statement $\mathbf{(x-y)-(x+y)+2y=0}$ using rules from the environment. Although simple, this statement already requires 22 proof steps and highlights the difficulty proving complex mathematical identities when using elementary proof steps.
In the rest of this appendix, we give more details about how we represent expressions in \eq, how we generate random theorems to provide initial training data, the list of rules we provided to the environment, and the set of expressions we use to evaluate the model.

\begin{figure}[h!]
\centering
\begin{align*}
\text{Statement to prove}&\hphantom{=0}\quad\quad\quad\quad\quad\text{Rule used}\\
\midrule
(x-y)-(x+y)+2y                &= 0 \quad\quad A-B = A+(-B)         \\
(x-y)+(-(x+y))+2y             &= 0 \quad\quad -(A+B) = (-A)+(-B)        \\
(x-y)+((-x)+(-y))+2y          &= 0 \quad\quad A+(B+C) = A+B+C\\
(x-y)+(-x)+(-y)+2y            &= 0 \quad\quad A+(-B) = A-B\\
(x-y)+(-x)-y+2y               &= 0 \quad\quad A+(-B) = A-B\\
(x-y)-x-y+2y                  &= 0 \quad\quad \text{int}(a+b) = \text{int}(a)+\text{int}(b) \\
(x-y)-x-y+(1+1)\times y       &= 0 \quad\quad A\times B = B\times A \\
(x-y)-x-y+y\times(1+1)        &= 0 \quad\quad A\times(B+C) = A\times B+A \times C \\
(x-y)-x-y+y\times 1+y\times 1 &= 0 \quad\quad A\times 1 = A         \\
(x-y)-x-y+y+y\times 1         &= 0 \quad\quad A-B = A+(-B)          \\
(x-y)-x+(-y)+y+y\times 1      &= 0 \quad\quad A+B = B+A             \\
(x-y)-x+y+(-y)+y\times 1      &= 0 \quad\quad A+(-B) = A-B             \\
(x-y)-x+y-y+y\times 1         &= 0 \quad\quad A-A = 0             \\
(x-y)-x+0+y\times 1           &= 0 \quad\quad A+0 = A               \\
(x-y)-x+y\times 1             &= 0 \quad\quad A-B = A+(-B)              \\
x+(-y)-x+y\times 1            &= 0 \quad\quad A+B = B+A           \\
(-y)+x-x+y\times 1            &= 0 \quad\quad A-A = 0             \\
(-y)+0+y\times 1              &= 0 \quad\quad A+0 = A               \\
(-y)+y\times 1                &= 0 \quad\quad A+B = B+A               \\
y\times 1+(-y)                &= 0 \quad\quad A+(-B) = A-B             \\
y\times 1-y                   &= 0 \quad\quad A\times 1 = A          \\
y-y                           &= 0 \quad\quad A-A = 0         \\
0                             &= 0                            \\
\end{align*}
\caption{\label{fig:simple_stuff_long} \textbf{Proof of the identity $\mathbf{(x-y)-(x+y)+2y=0}$ with elementary rules.} In this example, we provide at each step the current goal and the rule that is used to obtain the next goal. This example shows how difficult it can be to prove even simple statements in \eq as they may require a significant number of proof steps (22 in that case). This explains that proving more involved statements from \identities such as $\cosh(3x) = 4 \cosh(x)^3 - 3 \cosh(x)$ or even $\sin(2\pi + x) = \sin(x)$ can require to generate very large proof trees.
}
\end{figure}

\subsection{True expressions and numerical evaluation}
\label{sec:eq_verif_numeric} % referenced in next section 

Some theorems are trivial, either because their statements match the pattern of an assertion rule that has no assumptions (e.g. \smash{$x^{2}\geq 0$} or \smash{$e^{y-x} \neq 0$}), or because they do not contain any variable and an exact numerical evaluation can attest that they are true 
(e.g \smash{$(-1)/2<6$} or \smash{$1-7/4=-6/8$}).

To prevent the model from wasting budget in ``uninteresting'' branches, we automatically discard generated subgoals that can be trivially verified.
However, we only perform numerical verification of expressions without variables when they exclusively involve rational numbers. For instance, we will automatically close subgoals such as \smash{$5 < (-3)^2$} or \smash{$\frac12 > \frac14$}, but not \smash{$e^1 < e^2$} or \smash{$\cos(3) \neq 0$}. To prove that \smash{$e^1 < e^2$} the model will need to use, for instance, an assertion rule such as \smash{$A < B \implies e^A < e^B$} (\smash{$1 < 2$} will then be closed automatically).

\subsection{Environment vulnerabilities due to initial numerical approximations}
\label{sec:env_vulnerabilities} % referenced

In early implementations of the \eq environment, we found that the model was able to leverage vulnerabilities in the environment to reach a 100\% accuracy and to prove any statement. These issues where coming from numerical approximations that were initially allowed during the numerical verification of constant expressions~(c.f. Section~\ref{sec:eq_verif_numeric}).
To prevent these vulnerabilities, we restricted the numerical verification to rational expressions, in order to have an exact numerical evaluation and to avoid errors due to approximations. We give two examples of vulnerabilities found by the model when expressions were verified with an approximate numerical evaluation.

In Figure~\ref{fig:false_proof_exp}, the model manages to prove that $2 = 3$ by using the injectivity of the exponential function, and the fact that for NumPy, $\exp(-\exp(\exp(2))) = \exp(-\exp(\exp(3)))$. Evaluating the left and the right-hand side both numerically evaluate to $0.0$, and the environment incorrectly considered the expression to be valid.

In Figure~\ref{fig:false_proof_cos}, the model manages to prove that $0 \neq 0$ by first proving that \smash{$\cos (\pi/2) \neq 0$}, and combining this result with the fact that \smash{$\cos (\pi/2) = 0$}.
The imprecision came from the NumPy approximation of $\cos (\pi/2)$ in $6.123 \times 10^{-17}$, and in particular the fact that \smash{$(((\cos(\pi/2)^{0.5})^{0.5})^{0.5}) \approx 9.4 \times 10^{-3}$}, which was considered large enough by our threshold to be considered non-zero. By using this approximation, and the assertion rule \smash{$\sqrt{A} \neq 0 \implies A \neq 0$}, the model was able to conclude that \smash{$(((\cos(\pi/2)^{0.5})^{0.5})^{0.5}) \neq 0 \implies \cos (\pi/2) \neq 0 \implies 0 \neq 0$}.

\begin{figure}[h!]
\centering
\begin{alignat*}{5}
&&2 &= 3 \;\;                                    && \text{Statement to prove} \\
&\iff&   \; e^{2} &= e^{3}\;\;                   && \text{Rule: } A = B \iff e^A = e^B, \\
&\iff&   \; e^{e^{2}} &= e^{e^{3}}\;\;           && \text{Rule: } A = B \iff e^A = e^B, \\
&\iff&   \; -e^{e^{2}} &= -e^{e^{3}}\;\;         && \text{Rule: } A = B \iff -A = -B, \\
&\iff&   \; e^{-e^{e^{2}}} &= e^{-e^{e^{3}}}\;\; && \text{Rule: } A = B \iff e^A = e^B, \\
&\iff&   \; 0 &= 0\;\;                           && \text{Numerical evaluation} \\
\end{alignat*}
\caption{\textbf{False ``proof'' of $\mathbf{2=3}$ found by the model when allowing numerical approximation to verify constant expressions.} The model noticed that \smash{$\exp(-e^{e^{2}})$ = $\exp(-e^{e^{3}})$} is considered true by NumPy (as the left and the right hand side are both approximated to $0.0$) to conclude that $2=3$ using the injectivity of the exponential function.
}
\label{fig:false_proof_exp} % referenced
\end{figure}

\begin{figure}[h!]
\centering
\begin{alignat*}{2}
    0 \neq 0 &\iff \cos{\frac{\pi}{2}} \neq 0 &&\iff \left( \cos{\frac{\pi}{2}} \right)^{0.5} \neq 0  \\
    \iff \left(\left( \cos{\frac{\pi}{2}} \right)^{0.5}\right)^{0.5} \neq 0 &\iff \quad~\dots &&\iff \left(\left(\left( \cos{\frac{\pi}{2}} \right)^{0.5}\right)^{\dots}\right)^{0.5} \neq 0
\end{alignat*}

\caption{\textbf{False ``proof'' that $\mathbf{0 \neq 0}$ found by the model when allowing numerical approximation to verify constant expressions.}
Since $\cos(\frac{\pi}{2})$ evaluates to $6.123 \times 10^{-17}$ in NumPy (and not exactly to $0$), the model found that for any tolerance threshold applying the assertion rule \smash{$\sqrt{A} \neq 0 \implies A \neq 0$} enough times lead to an expression where the left-hand side evaluates numerically to a strictly positive value.
In particular, \smash{$(((\cos(\frac{\pi}{2})^{2^{-3}}) \approx 9.4 \times 10^{-3}$}, which was considered large enough by our threshold to be considered non-zero.
After that, any expressions $A$ and $B$ can be shown to be equal using the assertion rule \smash{$(A \times C = B \times C\; \wedge \;C \neq 0) \implies A = B$} where $C$ is chosen to be $0$ since $0\neq0$.
}
\label{fig:false_proof_cos} % referenced
\end{figure}

\newpage

\subsection{Random theorem generator}
\label{appendix:eq_gen} % referenced

While \mm and \lean come with a collection of annotated theorems that can be used for training, \eq does not have an equivalent of manually proved statements.
Instead, we generate a supervised training set of theorems to pretrain the model before we start the online training. We propose two simple generation procedures: a random walk, and a graph generation approach.

\paragraph{Random walk generation}

The random walk is the simplest way to generate a theorem. We start from an initial expression $A_0$ and a set of initial hypotheses, both randomly generated following the method of~\citet{Lample2020Deep}.
Then, we randomly apply an admissible transformation rule on $A_0$ to get an equivalent expression $A_1$.
The process is repeated, to get a sequence $A_0, A_1, \dots, A_N$ of equivalent expressions. The final theorem consists in proving that $A_0 = A_N$, and the proof corresponds to the sequence of rules sequentially applied.
To increase the diversity of generations, and to avoid sampling only rules without or with simple assumptions, we add a bias in the random sampling of rules to over-sample the underrepresented ones.

\paragraph{Graph generation}
Because of the simplicity of the random walk approach, the generated theorems usually tend to be very easy to prove, and the model quickly reaches a perfect accuracy on the generated theorems. Moreover, proofs generated by the random walk are only composed of transformation rules.
To generate a more diverse set of theorems, we also use a graph generation procedure, that creates a large acyclic graph of theorems, where each node is connected to its children by a rule in the environment.
To create such a graph, we proceed as follows. We first generate a set of initial hypotheses, and initialize the graph with a node for each hypothesis. We then randomly apply a transformation or assertion rule on nodes already in the graph.

For instance, if $A \leq B$ and $B \leq C$ are two nodes in the graph, then we can add the node $A \leq C$ using the assertion rule $A \leq B \wedge B \leq C \implies A \leq C$. If $x = y \times (z - 1)$ is a node in the graph, we can use the transformation rule $B \neq 0 \implies A / B = C \iff A = B \times C$ to add the node $x / y = z - 1$, provided that the node $y \neq 0$ is also in the graph. Required hypotheses that are trivially verifiable (e.g. $2 > 0$ or $e^{-x} > 0$) are automatically added to the graph.

\subsection{Translating \eq theorems to \lean}
\label{sec:eq_to_lean} % referenced

\paragraph{Exporting theorems to \lean.}
To enrich the existing \lean supervised dataset with synthetic data, we built a translator from \eq to \lean.
Although \eq statements are easy to translate, proofs can only be translated if they involve rules that also exist in \lean.
Since \eq is a modular environment where rules can be specified by the user, we created a collection of \eq rules from existing Mathlib statements.
Synthetic theorems can then be generated using the random walk or random graph approaches described in Section~\ref{appendix:eq_gen}, and converted into \lean to augment the existing supervised dataset. Examples of randomly generated \lean proofs are provided in Figure~\ref{proof:synthetic_examples}.

\begin{figure}[h!]
\begin{lstlisting}
theorem SYNTHETIC_0
  (x1 x3 x4 : ℝ) :
  ((0:ℝ) ≤ ((real.cos (real.cos ((-6:ℝ) / ((x1 - x4) / x3)))) / (2:ℝ))) :=
begin
  apply norm_num.nonneg_pos,
  apply half_pos,
  apply real.cos_pos_of_le_one,
  apply real.abs_cos_le_one,
end

theorem SYNTHETIC_1
  (x2 : ℝ)
  (h₀ : ((abs (5:ℝ)) < x2)) :
  ((1:ℝ) < (real.exp ((5:ℝ) + x2))) :=
begin
  rw real.one_lt_exp_iff,
  rw ← neg_lt_iff_pos_add,
  apply neg_lt_of_abs_lt h₀,
end

theorem SYNTHETIC_2
  (x1 x4 : ℝ)
  (h₀ : ((x4 * (real.exp x1)) < 10)) :
  ((-((abs ((x4 * (real.exp x1)) - 10)) / 2)) < ((abs (10 - (x4 * (real.exp x1)))) / 2)) :=
begin
  have h₁ : ((0:ℝ) < ((abs ((x4 * (real.exp x1)) - 10)) / 2)),
  apply half_pos,
  apply abs_pos_of_neg,
  apply sub_neg_of_lt h₀,
  apply norm_num.lt_neg_pos _ _ h₁,
  rw ← abs_sub_comm,
  apply half_pos,
  apply abs_pos_of_neg,
  apply sub_neg_of_lt h₀,
end
\end{lstlisting}
\caption{\textbf{Example of a randomly generated theorems in \lean}. The theorems were initially generated in the \eq environment using rules from the Mathlib library, and converted to \lean.}
\label{proof:synthetic_examples}
\end{figure}

\paragraph{Importing rules from Mathlib.}  To allow interfacing \eq and \lean, we automatically parsed Mathlib statements from the Lean library, and extracted theorems with a statement compatible with the \eq environment. Compatible theorems are converted into \eq transformation or assertion rules. Overall, we converted 1702 theorems from the \lean Library into our \eq environment. Details about the number of converted theorems are provided in Table~\ref{tab:converted_lean_rules}.

\begin{table}[h!]
\centering
\begin{tabular}{lccc}
\toprule
Rule type      & Natural numbers & Integers & Real numbers \\
\midrule
Transformation &  304 & 452 &  799 \\
Assertion      &  314 & 292 &  407 \\
\midrule
\textbf{Total} &  618 & 744 & 1206 \\
\bottomrule
\end{tabular}
\vspace{0.3cm}
\caption{\textbf{Number of \eq rules converted from \lean}. The converted \lean theorems can be used to generate synthetic theorems within the \eq environment. The generated theorems can then in turn be converted back to \lean, along with their proofs. Some theorems are generic and can be applied to different types of variables (e.g. \texttt{add\_comm}), and will appear in different categories. Overall, we automatically converted 1702 different \lean rules in our \eq environment.}
\label{tab:converted_lean_rules} % referenced
\end{table}

\subsection{Examples of identities solved by the model on \eq}
\label{sec:examples_identities_solved}

In this section, we give some examples of identities solved by the model. For each statement, we indicate the proof size and the proof depth, for the first proof found by the model, and for the optimal proof. We observe that the first proofs are sometimes very large, with more than 100 nodes, and that the model later manages to find shorter proofs as it improves.

\inserteqsolvedidentities

\newpage

\section{Proof Search Hyper-Parameters}
\label{appendix:hyperopt} % referenced

\looseness=-1 \algo depends on many hyper-parameters: the decoding hyper-parameters of the policy model and the search hyper-parameters. Selecting their optimal values would be difficult in practice, if not impractical, for several reasons.
First, the model is constantly evolving over time, and the optimal parameters may evolve as well.
For instance, if the model becomes too confident about its predictions, we may want to increase the decoding temperature to ensure a large diversity of tactics.
Second, even for a fixed model, the ideal parameters may be goal-specific. If an input statement can only be proved with very deep proofs, we will favor depth over breadth, and a small number of tactics per node. If the proof is expected to be shallow and to use rare tactics, we will want to penalize the exploration in depth and increase the number of tactics sampled per node.
Finally, there are simply too many parameters to tune and running each experiment is expensive.
Thus, we do not set \algo hyper-parameters to a fixed value, but sample them from pre-defined ranges at the beginning of each proof. 

The decoding parameters and the chosen distribution are the following:
\vspace{-0.3cm}
\begin{itemize}
\itemsep0em
    \item \textbf{Number of samples:} the number of tactics sampled from the policy model when a node is expanded. Distribution: uniform on discrete values $[8,16,32,48]$.
    \item \textbf{Temperature:} temperature used for decoding. Distribution: uniform on range $[0.8,2.0]$.
    \item \textbf{Length penalty:} length penalty used for decoding. Distribution: uniform on range $[0,1.2]$.
\end{itemize}

Also, for the search parameters we have:
\vspace{-0.3cm}
\begin{itemize}
\itemsep0em
\item \textbf{Number of expansions:} the search budget, i.e. the maximum number or nodes in the proof graph before we stop the search. Distribution: log-uniform with range $[1000,10000]$.
\item \textbf{Depth penalty:} an exponential value decay during the backup-phase, decaying with depth to favor breadth or depth. Distribution: uniform on discrete values $[0.8,0.9,0.95,1]$.
\item \textbf{Exploration:} the exploration constant $c$ in the policy (PUCT or RT). Distribution: log-uniform with range $[0.01,100]$.
\end{itemize}

\looseness=-1 When sampling proof search parameters during evaluation, we use the same distributions than at training time, with two differences: we fix the number of expansions to 5k in \lean and 10k in \mm.

\section{\mm \lean versions}
\label{appendix:hash_versions}

To compare our models in the same setup while working on this project, we ran all our experiments with a fixed version of \mm and \lean. In particular, all experiments were run with the following GitHub commits of \setmm, \lean, miniF2F, and Mathlib:
\begin{itemize}
    \item \url{https://github.com/metamath/set.mm}: 861bd3552636dcdb9cbc8df59d01b14520c72f82
    \item \url{https://github.com/leanprover/lean/}: tag/v3.3.0
    \item \url{https://github.com/openai/miniF2F}: 21723db70bbd030e034ed374db74cea4be1bf681
    \item \url{https://github.com/openai/miniF2F/tree/statement_curriculum_learning}: c9d827c871aff2ab0f5ec64a0d72e61111a7f072
    \item \url{https://github.com/leanprover-community/mathlib}: 9a8dcb9be408e7ae8af9f6832c08c021007f40ec
\end{itemize}

\newpage

\section{Time vs Depth minimization}
\label{appendix:time_vs_depth}

\begin{figure}[h]
\centering
\includegraphics[width=0.7\textwidth]{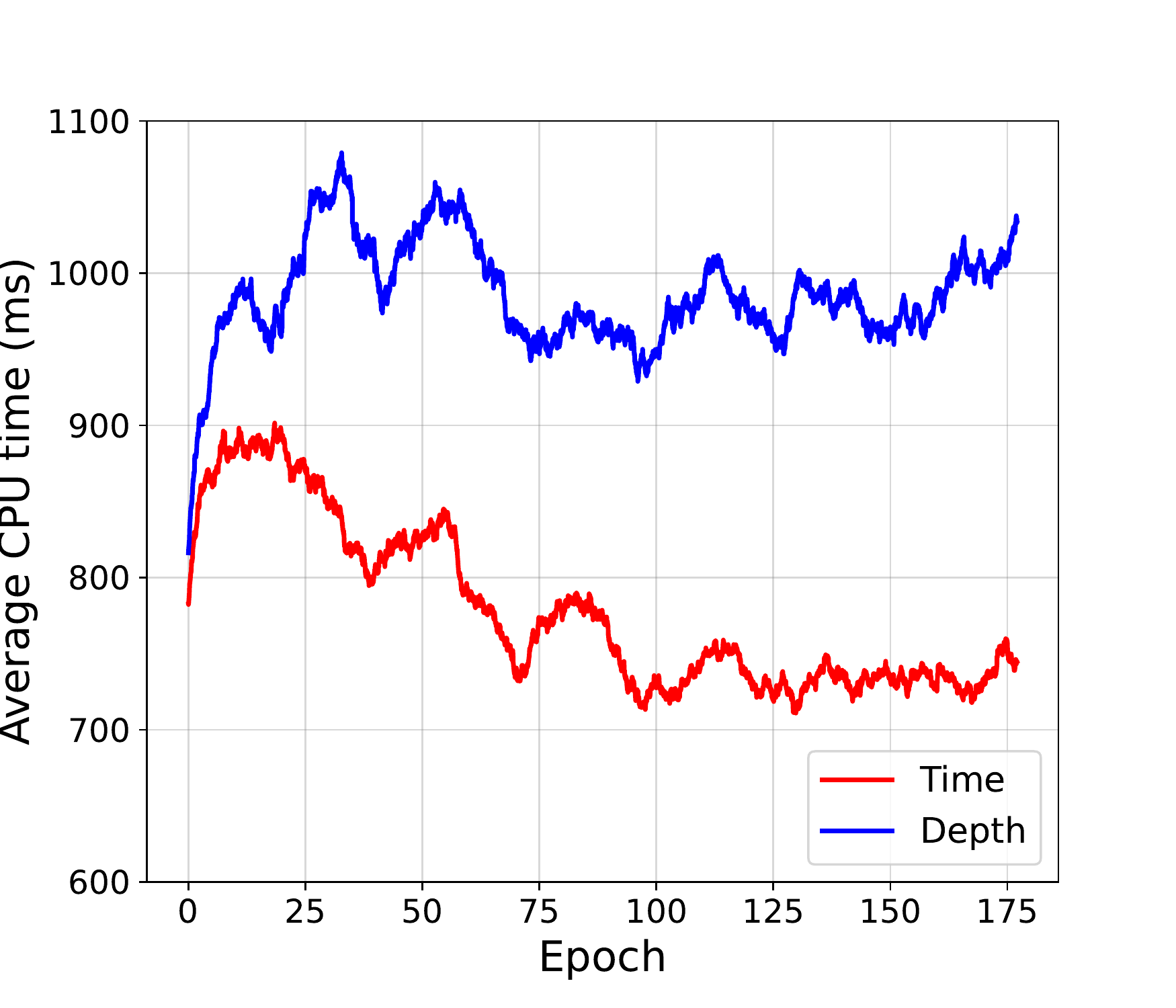}
\vspace{-0.3cm}
\caption{
\textbf{Average CPU time of proofs found by our model using different minimization objectives.} When the model is trained to generate proofs that minimize the running time in \lean (red), the average proof time is significantly lower.
\label{fig:lean_cpu} % referenced
}
\end{figure}

In \lean, powerful tactics like \texttt{ring}, \texttt{linarith}, or \texttt{simp} are helpful when it comes to proving theorems, but the model might become over-reliant on these automations and never learn what the underlying arguments of the proofs were. Since these tactics are usually slower than \texttt{rewrite} or \texttt{apply}, we experimented with minimizing the total \lean CPU time of the proof. After 3 days of training, the cumulative pass-rate is higher when minimizing this objective, with $50.4\%$ problems solved on \minivalid, compared to $47.5\%$ with the depth objective.
Besides, we found that a model trained to minimize the execution time will reduce the fraction of tactics that timeout by $40\%$.
In Table~\ref{tab:qual_timevsdepth} we provide a comparison of the tactic usage for two models trained with different minimization objectives, and we show the average \lean proof execution time in Figure~\ref{fig:lean_cpu}.

\begin{table}[h]
\centering
\begin{tabular}{lcc}
\toprule
          & Depth & Time \\
\midrule
ring      & 21    & 20   \\
norm\_num & 58    & 55   \\
linarith  & 66    & 66   \\
simp      & {\bf 62}    & 38   \\
\midrule
refl      & 3     & {\bf 7}   \\
exact     & 4     & {\bf 17}   \\
apply     & 11    & {\bf 35}   \\
rewrite   & 20    & {\bf 65}   \\
\bottomrule
\end{tabular}
\vspace{0.3cm}
\caption{\textbf{Qualitative comparison of the frequency of specific \lean tactics when using different minimization objective.} Tactics in the upper half are slow automated tactics, while tactics in the bottom half are faster and simpler.}
\label{tab:qual_timevsdepth}
\end{table}

\newpage

\section{Example Lean proofs}
\label{app:lean_proofs} % referenced.

In this section, we show examples of proofs found by our model.

\begin{figure}[h!]
\begin{lstlisting}
theorem imo_1964_p1_2 (n : %*$\mathbb{N}$*)) : %*$ \neg 7 | 2^n + 1$*) :=
begin
  rw nat.dvd_iff_mod_eq_zero,
  rewrite [nat.add_mod, nat.mod_eq_of_lt],
  obviously,
  apply nat.strong_induction_on n,
  induction n,
  {
    intros n IH,
    cases n,
    norm_num,
    cases n,
    norm_num,
    rw [nat.succ_eq_add_one, pow_succ],
    rw [nat.succ_eq_add_one, pow_succ],
    induction n,
    norm_num,
    rw [nat.succ_eq_add_one, pow_succ],
    norm_num [nat.mul_mod, ←mul_assoc],
    contrapose! IH,
    refine %*$\langle$*)n_n, nat.lt_succ_iff.mpr _, IH%*$\rangle$*),
    exact nat.le_succ_of_le (nat.le_succ _),
  },
  exact n_ih,
end
\end{lstlisting}
\caption{\small\textbf{A proof of the \texttt{imo\_1964\_p1\_2} problem found by our model.} The model shows that for any value of $n \in \mathbb{N}$, $2^n + 1$ is not divisible by $7$, by showing that $2^n \bmod 7 + 1 \neq 0$, and $2^n \bmod 7 + 1 < 7$. The second part of the proof uses strong induction and the fact that $2^n \equiv 2^{n+3} \mod 7$. We provide a version of the proof that was automatically cleaned by removing unnecessary tactics and tactic arguments.}
% \label{fig:my_label}
\end{figure}

\begin{figure}[h!]
\begin{lstlisting}
theorem imo_2001_p6
  (a b c d : ℕ)
  (h₀ : 0 < a ∧ 0 < b ∧ 0 < c ∧ 0 < d)
  (h₁ : d < c)
  (h₂ : c < b)
  (h₃ : b < a)
  (h₄ : a * c + b * d = (b + d + a - c) * (b + d - a + c)) :
  ¬ nat.prime (a * b + c * d) :=
begin
  contrapose h₄,
  rw mul_comm,
  simp [nat.prime, not_le_of_gt h₀.1, not_forall, not_le_of_gt h₃,
        nat.mul_sub_right_distrib, nat.add_comm],
  contrapose! h₄,
  contrapose! h₄,
  apply has_lt.lt.ne,
  apply nat.lt_sub_right_of_add_lt,
  nlinarith,
end
\end{lstlisting}
\caption{\small\textbf{A proof found by our model of another IMO problem in \miniff.} Although the proof is valid, the statement is erroneous. The hypothesis $h_4$ : $b + d - a + c$ actually represents $\max(b + d - a, 0) + c$. This is due to \lean's \texttt{nat} type behaviour where $(a:\mathbb{N})-(b:\mathbb{N})=(0:\mathbb{N})$ if $b \geq a$. This makes the exercise easier than it should be, and the proof is no longer valid on the fixed statement.}
\end{figure}

\end{document}